\documentclass[journal]{IEEEtran}

\usepackage{amsmath,graphicx}
\usepackage{balance}
\usepackage{amsmath}
\usepackage{amsfonts}
\usepackage{graphicx}
\usepackage{pdfpages}
\usepackage{cite}
\usepackage{subfig} 
\usepackage[T1]{fontenc} 
\usepackage{array}
\usepackage{url}
\usepackage{float}

\usepackage{color}
\usepackage{wrapfig}
\usepackage{lipsum}
\usepackage[hidelinks]{hyperref}
\usepackage{multirow}
\usepackage{color}
\usepackage{tabularx}
\usepackage{gensymb}
\newcolumntype{L}{>{\arraybackslash}m{4.5cm}}  
\newcolumntype{M}{>{\arraybackslash}m{5cm}}  
\begin{document}
\title{Machine Learning Techniques and Applications \\For Ground-based Image Analysis}
\author{Soumyabrata~Dev,~\IEEEmembership{Student Member,~IEEE,}
        Bihan~Wen,~\IEEEmembership{Student Member,~IEEE,}\\
        Yee~Hui~Lee,~\IEEEmembership{Senior~Member, IEEE,}
        and Stefan~Winkler,~\IEEEmembership{Senior Member,~IEEE}
\thanks{Manuscript received 15-Dec-2014; revised 23-Jun-2015; accepted 03-Dec-2015. This work is supported by a grant from Singapore's Defence Science \& Technology Agency (DSTA).}        
\thanks{S.\ Dev and Y.\ H.\ Lee are with the School of Electrical and Electronic Engineering, Nanyang Technological University, Singapore (e-mail: soumyabr001@e.ntu.edu.sg, EYHLee@ntu.edu.sg).}
\thanks{B.\ Wen is with the Advanced Digital Sciences Center (ADSC), the Department of Electrical and Computer Engineering and the Coordinated Science Laboratory, University of Illinois at Urbana-Champaign, IL 61801, USA (e-mail: \mbox{bwen3@illinois.edu}). }
\thanks{S.\ Winkler is with the Advanced Digital Sciences Center (ADSC), University of Illinois at Urbana-Champaign, Singapore (e-mail: \mbox{Stefan.Winkler@adsc.com.sg}).}
\thanks{Send correspondence to S.\ Winkler, E-mail: stefan.winkler@adsc.com.sg.}
}

\maketitle

\begin{abstract}
Ground-based whole sky cameras have opened up new opportunities for monitoring the earth's atmosphere. These cameras are an important complement to satellite images by providing geoscientists with cheaper, faster, and more localized data. The images captured by whole sky imagers can have high spatial \emph{and} temporal resolution, which is an important pre-requisite for applications such as solar energy modeling, cloud attenuation analysis, local weather prediction, etc. 

Extracting valuable information from the huge amount of image data by detecting and analyzing the various entities in these images is challenging. However, powerful machine learning techniques have become available to aid with the image analysis. This article provides a detailed walk-through of recent developments in these techniques and their applications in ground-based imaging. We aim to bridge the gap between computer vision and remote sensing with the help of illustrative examples. We demonstrate the advantages of using machine learning techniques in ground-based image analysis via three primary applications -- segmentation, classification, and denoising.
\end{abstract}


\IEEEpeerreviewmaketitle

\section{Introduction}

\IEEEPARstart{S}{atellite} images are commonly used to monitor the earth and analyze its various properties. They provide remote sensing analysts with accurate information about various earth events. Satellite images are available in different spatial and temporal resolutions and also across various ranges of the electromagnetic spectrum, including visible, near- and far-infrared regions. For example, multi-temporal satellite images are extensively used for monitoring forest canopy changes~\cite{forest} or evaluating sea ice concentrations~\cite{ice2008}. 

The presence of clouds plays a very important role in the analysis of satellite images. NASA's Ice, Cloud, and land Elevation Satellite (ICESat) has demonstrated that $70\%$ of the world's atmosphere is covered with clouds~\cite{NASA-cloud}. Therefore, there has been renewed interest amongst the remote sensing community to further study clouds and their effects on the earth. 

Satellite images are a good starting point for monitoring the earth's atmosphere. However, they have either high temporal resolution (e.g.\ geostationary satellites) or high spatial resolution (e.g.\ low-orbit satellites), but never both. In many applications like solar energy production~\cite{solar_irr_pred}, local weather prediction, tracking contrails at high altitudes~\cite{contrail_WSI}, studying aerosol properties~\cite{fuse_AOD}, attenuation of communication signals \cite{cloud_model_compare,radiosonde2014}, we need data with high spatial \emph{and} temporal resolution. This is why ground-based sky imagers have become popular and are now widely used in these and other applications. The ready availability of high-resolution cameras at a low cost facilitated the development of various models of sky imagers. 

A Whole Sky Imager (WSI) consists of an imaging system placed inside a weather-proof enclosure that captures the sky at user-defined intervals. A number of WSI models have been developed over the years. A commercial WSI (\mbox{TSI-440}, \mbox{TSI-880}) manufactured by Yankee Environmental Systems (YES) is used by many researchers~\cite{Long,Souza,Ghonima2012}.  Owing to the high cost and limited flexibility of commercial sky imagers, many research groups have built their own WSI models \cite{Sylvio,sky_imager2008,Kazantzidis2012,synerg,orograph2007,uwe2000,infrared_UK,how_weather}. For example, the Scripps Institution of Oceanography at the University of California San Diego has been developing and using WSIs as part of their work for many years~\cite{WSI_UCSD}.  Similarly, our group designed the Wide-Angle High-Resolution Sky Imaging System (WAHRSIS) for cloud monitoring purposes \cite{WAHRSIS,IGARSS2015,IGARSS2015c}. Table~\ref{tab:diff-database} provides an overview of the types of ground-based sky cameras used by various organizations around the world, and their primary applications. 

\begin{table*}[htb]
\footnotesize
\centering
\begin{tabular}{ l|l|l|l }
\hline 
\textbf{Application}  & \textbf{Organization} & \textbf{Country} &  \textbf{WSI Model}  \\
\hline
Air traffic control~\cite{infrared_UK} & Campbell Scientific Ltd. & United Kingdom & IR NEC TS9230  \\   
Cloud attenuation~\cite{WAHRSIS,IGARSS2015,IGARSS2015c} & Nanyang Technological University Singapore & Singapore & WAHRSIS \\  
Cloud characterization~\cite{sky_imager2008}& Atmospheric Physics Group & Spain & GFAT All-sky imager\\ 
Cloud classification~\cite{Sylvio} & Brazilian Institute for Space Research & Brazil & TSI-440\\
Cloud classification~\cite{Kazantzidis2012} & Laboratory of Atmospheric Physics & Greece & Canon IXUS II with FOV $180^{\circ}$   \\ 
Cloud macrophysical properties~\cite{Long} & Pacific Northwest National Laboratory & United States & Hemispheric Sky Imager\\ 
Cloud track wind data monitoring~\cite{synerg} & Laboratoire de M\'et\'eorologie Dynamique & France & Nikon D100 with FOV $63^{\circ}$\\ 
Convection~\cite{orograph2007}& Creighton University & United States & Digital Camera   \\ 
Radiation balance~\cite{uwe2000}& Lindenberg Meteorological Observatory & Germany & VIS/NIR 7   \\  
Solar power forecasting~\cite{Ghonima2012}& Solar Resource Assessment \& Forecasting Laboratory & United States & TSI-440 \\ 
Weather monitoring~\cite{TSI880} & Pacific Northwest National Laboratory & United States & TSI-880\\ 
Weather reporting~\cite{how_weather}& Ecole Polytechnique F\'ed\'erale de Lausanne & Switzerland & Panorama Camera\\ 
\hline  
\end{tabular}
\caption{Overview of various ground-based whole sky imagers and their intended applications.}
\label{tab:diff-database}
\end{table*}

\section{Machine Learning for Remote Sensing Data} \label{sec2}

The rapid increase in computing power has enabled the use of powerful machine learning algorithms on large datasets.  Remote sensing data fill this description and are typically available in different temporal, spatial, and spectral resolutions. For aerial surveillance and other monitoring purposes, RGB images are captured by low-flying aircraft or drones. Multispectral data are used for forest, land, and sea monitoring. Quite recently, hyperspectral imaging systems with very narrow bands are employed for identifying specific spectral signatures for  agriculture and surveillance applications. 

In cloud analysis, one example of such remote sensing data are ground-based images captured by WSIs. With these images, one can monitor the cloud movement and predict the clouds' future location, detect and track contrails and monitor aerosols. This is important in applications such as cloud attenuation, solar radiation modeling etc., which require high temporal and spatial resolution data. The requirement for high-resolution data is further exemplified by places where weather conditions are more localized. Such microclimates are prevalent mainly near bodies of water which may cool the local atmosphere, or in heavily urban areas where buildings and roads absorb the sun's energy (Singapore, the authors' home, being a prime example of such conditions). This leads to quicker cloud formation, which can have sudden impact on signal attenuation or solar radiation. Therefore, high-resolution ground-based imagers are required for a continuous and effective monitoring of the earth's atmosphere.

\begin{figure}[htb]
\centering
   \includegraphics[width=\columnwidth]{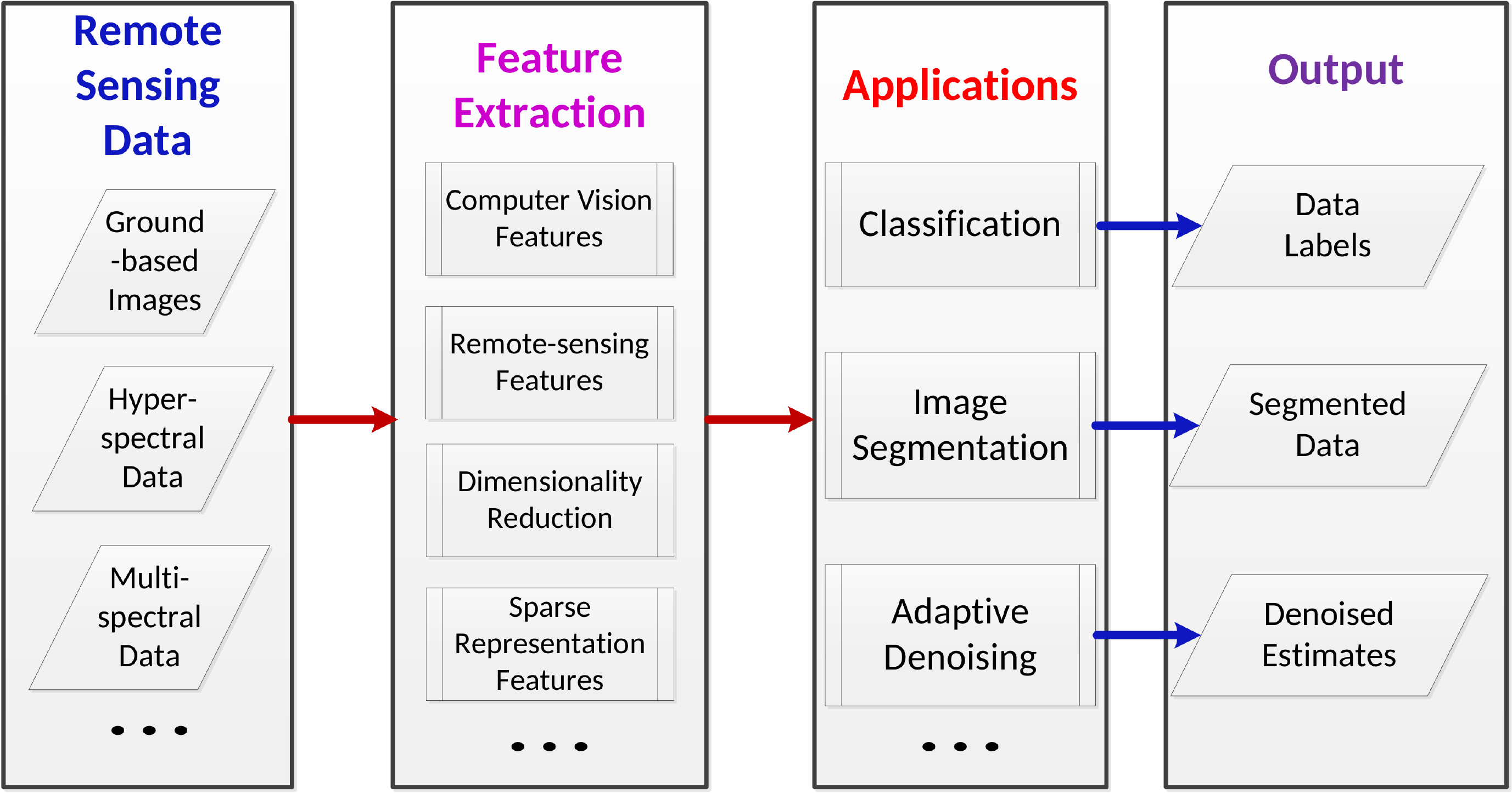}
\caption{High-level schematic framework of remote sensing data analysis with machine learning techniques. }\label{fig:block_d}
\end{figure}

In this paper, we show how a number of popular state-of-the-art machine learning methods can be effectively used in remote sensing in general and ground based image analysis in particular.
A high-level schematic framework for this is shown in Fig.~\ref{fig:block_d}.

There are a number of challenges with applying machine learning techniques in remote sensing. While the high dimensionality of remote sensing data can provide rich information and a complex data model, it is normally expensive and difficult to create a sufficient amount of labeled data for reliable supervised training. Additionally, the influence of atmospheric noise and interference introduces error and variance in the acquired training data. Thus, without effective regularization and feature extraction, overfitting can occur in the learned model, which may eventually affect the performance of the method. 

Moreover, processing the rich amount of high-dimensional data directly leads to high computational cost and memory requirements, while the large amount of data redundancy fails to facilitate the learning significantly. Therefore, appropriate feature extraction is crucial in machine learning, especially for remote sensing applications. In Section~\ref{sec3}, we discuss some of the most popular types of features, including computer vision features, remote-sensing features, dimensionality reduction, and sparse representation features. Instead of the full-dimensional raw input data, these extracted features are used for subsequent analysis in different application domains. Illustrative examples are also provided for these types of features to demonstrate their utility and effectiveness. 

Using  three primary applications as examples, namely segmentation, classification and denoising, we show in Section~\ref{sec4} that a learning-based framework can potentially perform better compared to heuristic approaches. Image segmentation is the task of categorizing pixels into meaningful regions, which share similar properties, belong to same group, or form certain objects. Classification is the problem of recognizing objects based on some pre-defined categories.  Denoising estimates the true signals from their corrupted observations. 

In this paper, we show how a number of popular state-of-the-art machine learning methods can be effectively used in remote sensing in general and ground based image analysis in particular.
A high-level schematic framework for this is shown in Fig.~\ref{fig:block_d}. 

\section{Feature Extraction} \label{sec3}
\label{sec:feat-ext}
Effective image features are important for computational efficiency and enhanced performance in different applications. Because of the high dimensionality of the data, it is difficult and inefficient to learn from the raw data directly. Moreover, the effect of collinearity amongst the input variables and the presence of noise degrade the performance of the algorithms to a great extent. Therefore, discriminative features should be chosen carefully from the input data. 

It is beyond the scope of this tutorial to encompass and list all existing feature extraction techniques. We focus on those popular feature extractors that are widely used in the remote sensing community, and that show promise for ground-based image analysis. Based on the application domains and the nature of the techniques, we distinguish four primary categories of feature extraction techniques in this paper, which will be discussed in more detail below:
\begin{itemize}
\item Computer vision features;
\item Remote-sensing features;
\item Dimensionality reduction;
\item Sparse representation features.
\end{itemize}

\subsection{Computer Vision Features}
Traditional computer vision feature extraction techniques mainly consist of corner and edge detectors. The term corner has varied interpretations. Essentially, a corner denotes a region where there is a sharp variation in brightness. These corner points may not always represent the projection of a 3D corner point in the image. In an ideal scenario, the feature detector should detect the same set of corners under any affine transformation of the input images. 

The most commonly used algorithm is the Harris corner detector \cite{Harris}. It relies on a small window that slides across the image and looks for variations of intensity changes. In  automatic satellite image registration, Harris corner detection has been used to extract feature points from buildings and natural terrain \cite{Misra2012,Ying2014}, for example. 

Aside from corners, blobs are also popular discriminatory features. Blobs are small image regions that possess similar characteristics with respect to color, intensity etc. Popular blob detectors are Difference of Gaussians (DoG), Scale-Invariant Feature Transform (SIFT) \cite{SIFT} and Speeded-Up Robust Features (SURF) \cite{SURF}. These feature descriptors have high invariability to affine transformations such as rotation. 

DoG is a band-pass filter that involves the subtraction of two blurred versions of the input image. These blurred versions are obtained by convolving the image with two Gaussian filters of different standard deviations. Because of its attractive property to enhance information at certain frequency ranges, DoG can be used to separate the specular reflection from Ground Penetrating Radar (GPR) images~\cite{DOG-landmine}. This is necessary for the detection of landmines using radar images. DoG also has wide applications in obtaining pan-sharpened images, which have high spectral and spatial resolutions~\cite{DOG-pan}.

SIFT and SURF are two other very popular blob-based feature extraction techniques in computer vision that are widely used in remote sensing analysis. SIFT extracts a set of feature vectors from an image that are invariant to rotation, scaling, and translation. They are obtained by detecting extrema in a series of sampled and smoothed versions of the input image. It is mainly applied to the task of image registration in optical remote sensing images~\cite{Seda-SIFT} and multispectral images~\cite{Li-SIFT}.  Unlike SIFT, SURF uses integral images to detect feature points in the input image. Its main advantage is its faster execution as compared to SIFT. Image matching on Quickbird images is done using SURF features \cite{Wu2012}; Song et al.\  \cite{Song2010} proposed a robust retrofitted SURF algorithm for remote sensing image registration.

These corner and blob detectors are essentially local features, i.e.\ they have a spatial interpretation, exhibiting similar properties of color, texture, position, etc.\ in their neighborhood~\cite{Blobworld_PAMI}. These local features help retain the local information of the image, and provide cues for  applications such as image retrieval and image mining. 

In addition to corner and blob detectors, local features based on image segmentation are also popular. The entire image is divided into several sub-images, by considering the boundaries between different objects in the image. The purpose of segmentation-based features is to find homogeneous regions of the image, which can  subsequently be used in an image segmentation framework.

Pixel-grouping techniques group pixels with similar appearance. Popular approaches such as the superpixel method \cite{Ren2003} have also been applied for remote sensing image classification. Recently Vargas et al.\ \cite{Vargas2015} presented a Bag-Of-Words (BoW) model using superpixels for multispectral image classification. Zhang et al.\ \cite{Zhang2012} use superpixel-based feature extraction in aerial image classification. 

Another popular technique of pixel-grouping is graph-based image representation, where pixels with similar properties are connected by edges. Graph-theoretic models allow for encoding the local segmentation cues in an elegant and systematic framework of nodes and edges. The segmented image is obtained by cutting the graph into sub-graphs, such that the similarity of pixels within a sub-graph is maximized. A good review of the various graph-theoretical models in computer vision is provided by Shokoufandeh and Dickinson \cite{review_GT}.

\begin{figure*}[tb]
\centering
   \subfloat[]{\includegraphics[height=1.5in]{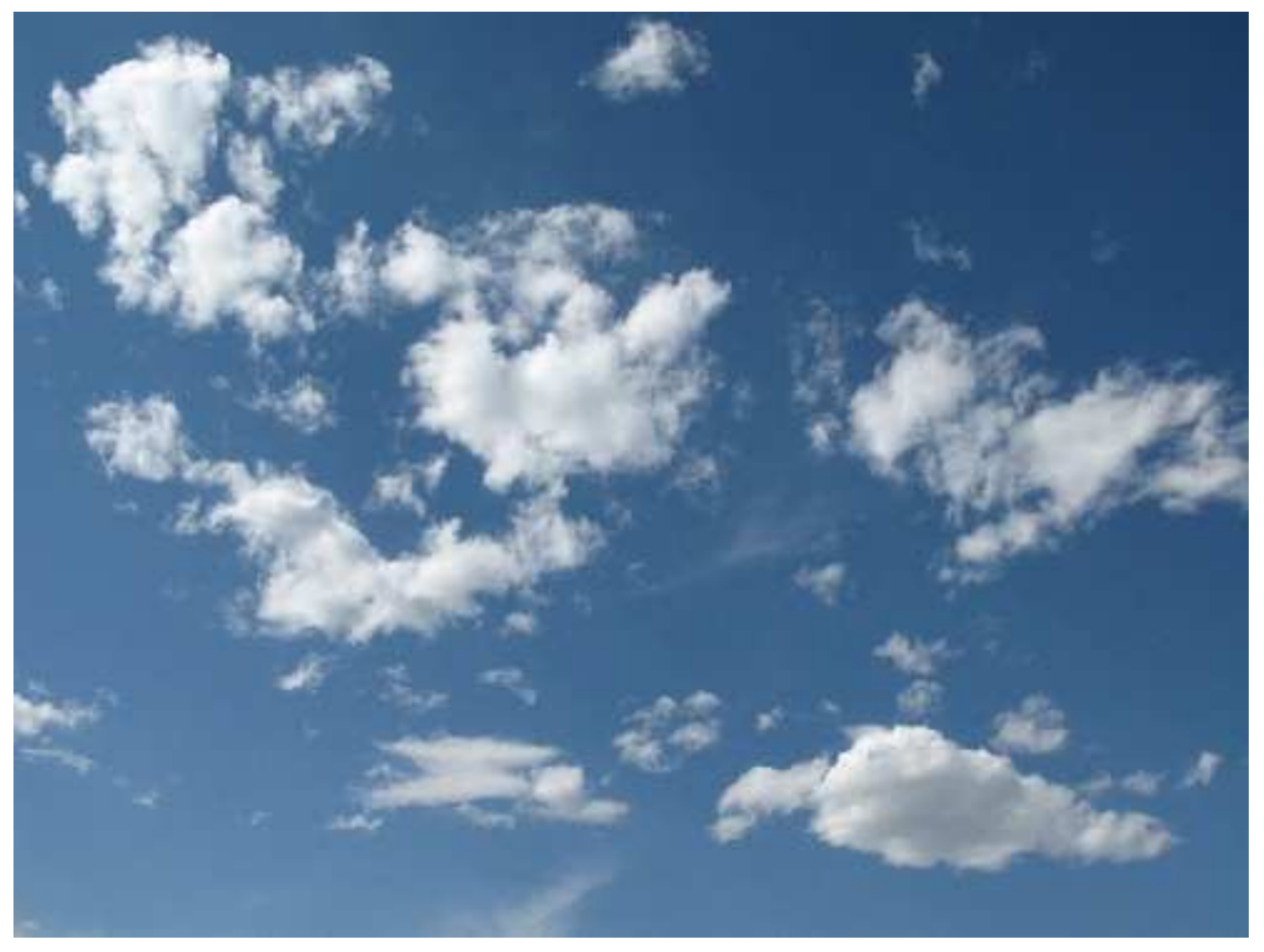}} 	
   \subfloat[]{\includegraphics[height=1.5in]{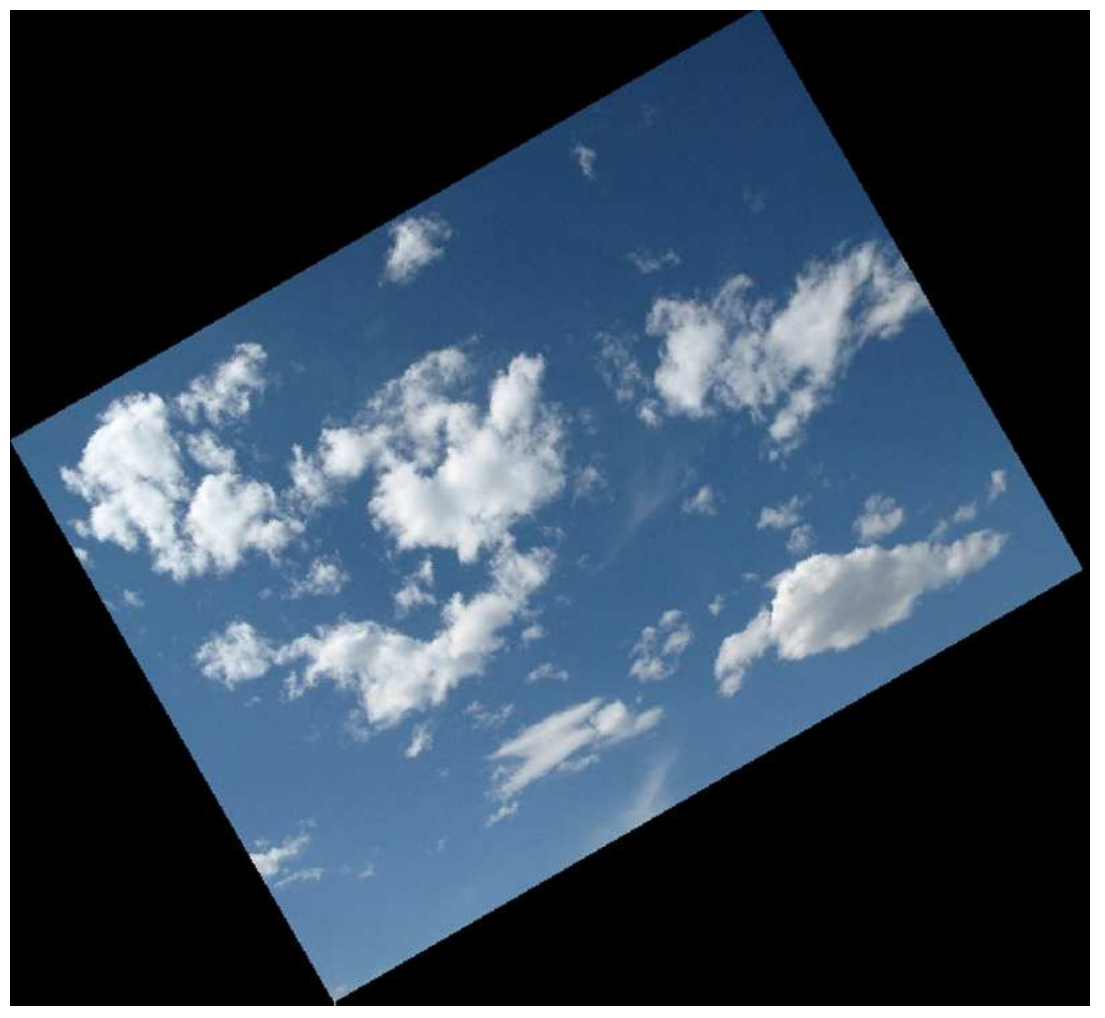}} 
   \subfloat[]{\includegraphics[height=1.5in]{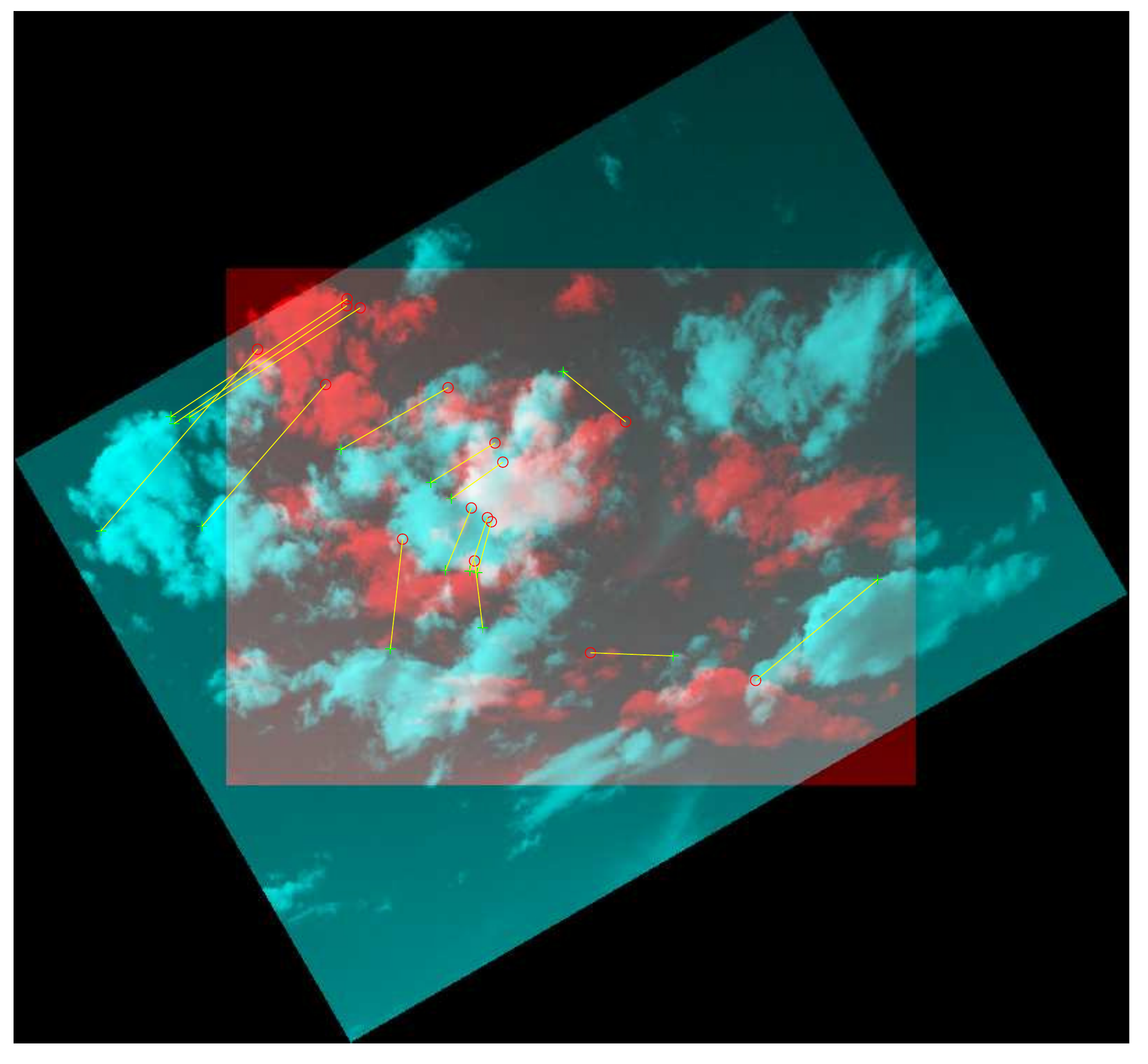}}  
   \subfloat[]{\includegraphics[height=1.5in]{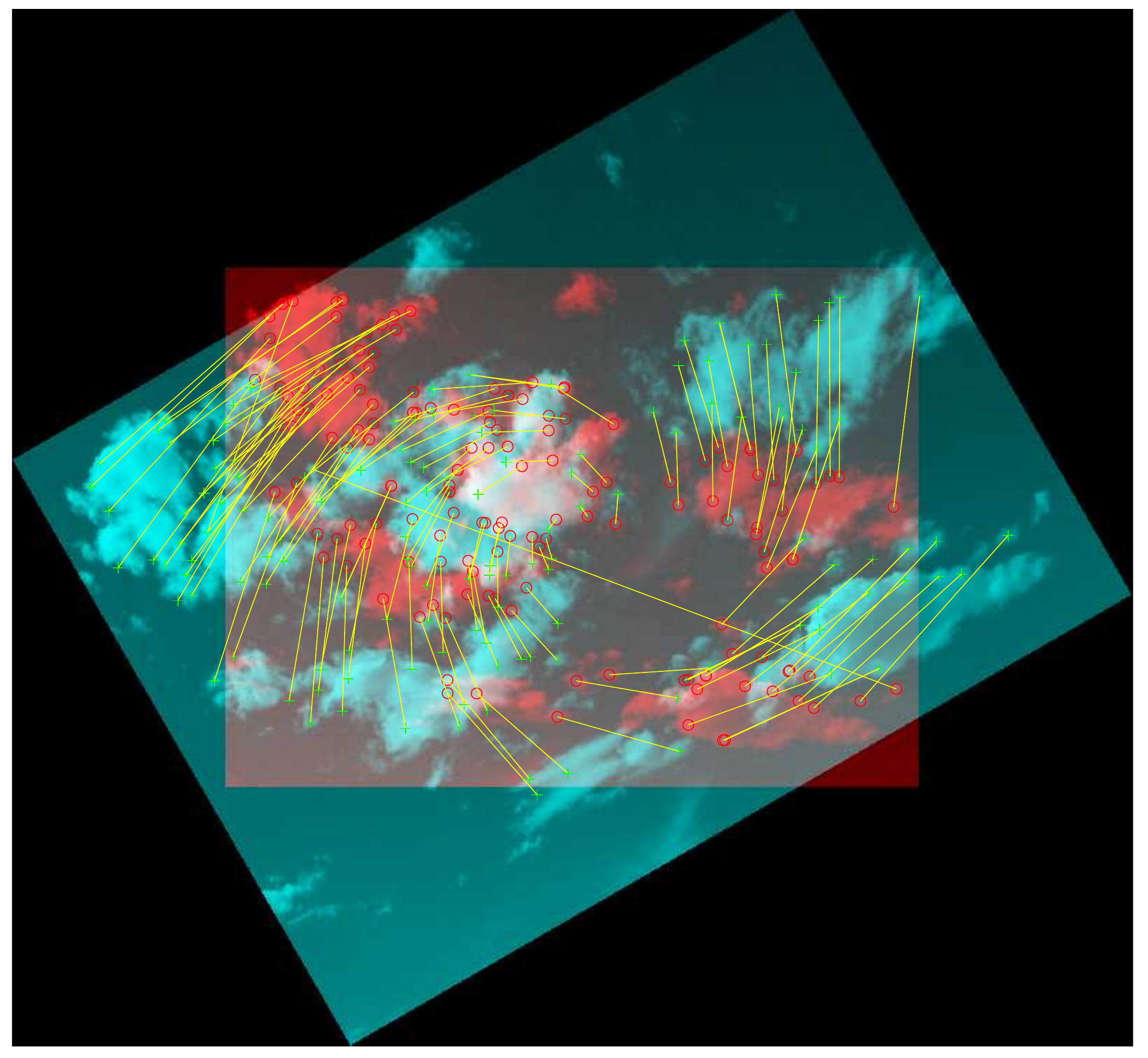}} 
\caption{Feature matching between (a) original image and (b) transformed image (scaled by a factor of 1.3 and rotated by $30^{\circ}$). (c) Candidate matches using Harris corner detector. (d) Candidate matches using SURF detector.}
\label{fig:feat-match}
\end{figure*}

In order to illustrate the corner and blob detector features in the context of ground-based image analysis, we provide an illustrative example by considering a sample image from the HYTA database~\cite{Li2011}. The original image is scaled by a factor of $1.3$ and rotated by $30^{\circ}$. Figure~\ref{fig:feat-match} shows candidate matches between input and  transformed image using the Harris corner detector  and SURF. As clouds do not possess strong edges, the number of detected feature points using the Harris corner detector is far lower than that of the SURF detector. Furthermore, the repeatability of the SURF detector is higher than the corner detector for the same amount of scaling and rotation. 

\subsection{Remote-sensing Features}
\label{sec:RS-feat}
In remote sensing, hand-crafted features exploiting the characteristics of the input data are widely used for image classification \cite{hand_crafted}. It involves the generation of a large number of features that capture the discriminating cues in the data. The user makes an educated guess about the most appropriate features. Unlike the popular computer vision feature extraction techniques presented above, remote sensing features use their inherent spectral and spatial characteristics to identify discriminating cues of the input data. They are not learning-based, but are derived empirically from the input data and achieve good results in certain applications. 

For example, Heinle et al.\ \cite{Heinle2010} proposed a 12-dimensional feature vector that captures color, edge, and texture information of a sky/cloud image; it is quite popular in cloud classification. The raw intensity values of RGB aerial images have also been used as input features~\cite{ISPRS12}. In satellite imagery, Normalized Difference Vegetation Index (NDVI) is used in association with the raw pixel intensity values for monitoring land-cover, road structures, and so on \cite{Bhandari2012}. In high-resolution aerial images, neighboring pixels are  considered for the generation of feature vectors. This results in the creation of e.g.\ $3 \times 3$, $15 \times 15$, $21 \times 21$ etc.\ pixel neighborhoods. Furthermore, in order to encode the textural features of the input images, Gabor- and edge-based texture filters are used, e.g.\ for aerial imagery \cite{Shao1994} or landscape image segmentation \cite{Galun2003}. Recently, we have used a modified set of Schmid filters for the task of cloud classification \cite{ICIP2015b}.

\subsection{Dimensionality Reduction}
\label{S:DR}
Remote sensing data are high-dimensional in nature. Therefore, it is advisable to reduce the inherent dimensionality of the data considerably, while capturing sufficient information in the reduced subspace for further data processing. In this section, we discuss several popular Dimensionality Reduction (DR) techniques and point to relevant remote sensing applications. A more detailed review of various DR techniques can be found in \cite{multivariate-tutorial}. 

Broadly speaking, DR techniques can be classified as either linear or non-linear. Linear DR methods represent the original data in a lower-dimensional subspace by a linear transformation, while non-linear methods consider the non-linear relationship between the original data and the features. In this paper, we focus on linear DR techniques because of their lower computational complexity and simple geometric interpretation; a brief overview of the different techniques is provided in Table~\ref{tabh5}. A detailed treatment of the various methods can be found in \cite{unified_linear}.  

\begin{table}[htb]
\centering
\begin{tabular}{clcc}
\hline
\textbf{Technique} & \textbf{Maximized Objectives} & \textbf{Supervised} & \textbf{Convex}\\ \hline
 PCA & Data variance & No & Yes \\ \hline	
  \multirow{2}{*}{FA} & Likelihood function of & \multirow{2}{*}{No} & \multirow{2}{*}{No} \\
            & underlying distribution parameters  &    &   \\ \hline
 \multirow{2}{*}{LDA} & Between-class variability over & \multirow{2}{*}{Yes} & \multirow{2}{*}{Yes} \\
            & within-class variability  &    &   \\ \hline
 \multirow{2}{*}{NCA} & Stochastic variant of the & \multirow{2}{*}{Yes} & \multirow{2}{*}{No} \\
            & LOO score  &   &   \\ \hline
\end{tabular}
\caption{Summary of linear dimensionality reduction techniques.}
\label{tabh5}
\end{table}

We denote the data as $\mathbf{X} = \begin{bmatrix}x_{1} \mid x_{2}\mid &...& \mid x_{n}\end{bmatrix} \in {\rm I\!\mathbf{R}}^{N \times n}$, where each $x_{i} \in {\rm I\!\mathbf{R}}^{N}$ represents a vectorized data point, $N$ denotes the data dimensionality, and $n$ is the data size. The corresponding features are denoted as $\mathbf{Z} = \begin{bmatrix}z_{1} \mid z_{2}\mid &...& \mid z_{n}\end{bmatrix} \in {\rm I\!\mathbf{R}}^{K \times n}$, where each $z_{i} \in {\rm I\!\mathbf{R}}^{K}$ is the feature representation of $x_{i}$, and $K$ denotes the feature dimensionality. 

\textbf{Principal Component Analysis (PCA)} is one of the most common and widely used DR techniques. 
It projects the $N$-dimensional data $\mathbf{X}$ onto a lower $K$-dimensional (i.e., $K \le N$) feature space as $\mathbf{Z}$ by maximizing the captured data variance, or equivalently, minimizing the reconstruction error. PCA can be represented as:
\begin{equation}
\label{eq:PCA}
\mathbf{Z} = \mathbf{U}^{T}\mathbf{X},
\end{equation}
where $\mathbf{U} \in {\rm I\!\mathbf{R}}^{N \times K}$ is formed by the principal components, which are orthonormal and can be obtained from the eigenvalue decomposition of the data covariance matrix. The objective function is convex, thus convergence and global optimality are guaranteed. In the field of remote sensing, PCA is often used to reduce the number of bands in multispectral and hyperspectral data. It is also widely used for change detection in forest fires and land-cover studies. Munyati \cite{PCA_wetland} used PCA as a change detection technique in inland wetland systems using Landsat images, observing that most of the variance was captured in the near-infrared reflectance. Subsequently, the image composite obtained from the principal axes was used in change detection.

\textbf{Factor Analysis (FA)} is based on the assumption that the input data $\mathbf{X}$ can be explained by a set of underlying `factors'. These factors are relatively independent of each other and are used to approximately describe the original data. The input data $\mathbf{X}$ can be expressed as a linear combination of $K$ factors with small independent errors $\mathbf{E}$
\begin{equation}
\label{eq:FA}
\mathbf{X} = \sum_{i = 1}^{K} F_i Z_i + \mathbf{E},
\end{equation}
where $\left \{ F_i \right \}_{i=1}^{K} \in {\rm I\!\mathbf{R}}^{N}$ are the different derived factors, and $\mathbf{Z}_i$ denotes the $i^\mathrm{th}$ row of the feature matrix $\mathbf{Z}$. The error matrix $\mathbf{E}$ explains the variance that cannot be expressed by any of the underlying factors. The factors $\left \{ F_i \right \}_{i=1}^{K}$ can be found by maximizing the likelihood function of the underlying distribution parameters. To our knowledge, there is no algorithm with a closed-form solution to this problem. Thus expectation-maximization (EM) is normally used, but it offers no performance guarantee due to the non-convex problem formulation. In remote sensing, FA is used in aerial photography and ground surveys. Doerffer and Murphy \cite{tidal_FA} have used FA techniques in multispectral data to extract latent and meaningful within-pixel information.

Unlike PCA and FA, which are unsupervised (i.e., using unlabeled data only), \textbf{Linear Discriminant Analysis (LDA)} is a supervised learning technique that uses training data class labels to maximize class separability. Given all training data $\mathbf{X}$ from $p$ classes, the mean of the $j^\mathrm{th}$ class $C_j$ is denoted as $\mu_j$, and the overall mean is denoted as $\mu$. We define the within-class covariance matrix $\mathbf{S}_W$ as:
\begin{equation}
\label{eq:S_W}
\mathbf{S}_W = \sum_{j=1}^{p} \sum_{i \in C_j}(x_i - \mu_j)(x_i - \mu_j)^{T},
\end{equation}
and the between-class covariance matrix $\mathbf{S_B}$ as:
\begin{equation}
\label{eq:S_B}
\mathbf{S}_B = \sum_{j=1}^{p} (\mu_j - \mu)(\mu_j - \mu)^{T}.
\end{equation}
Thus, the maximum separability can be achieved by maximizing the between-class variability over within-class variability over the desired linear transform $W$ as
\begin{equation}
\label{eq:LDA}
\max_{\mathbf{W}}\: \frac{\mathrm{tr}\left\{\mathbf{W}\mathbf{S}_B\mathbf{W}^{T}\right\}}{\mathrm{tr}\left\{\mathbf{W}\mathbf{S}_W\mathbf{W}^{T}\right\}},
\end{equation}
where $\mathrm{tr}\left\{\cdot\right\}$ denotes the trace of the matrix. The solution provides the linear DR mapping $\mathbf{W}$ that is used to produce LDA feature $\mathbf{Z} = \mathbf{W} \mathbf{X}$. 

LDA is widely used for the classification of hyperspectral images. In such cases, the ratio of the number of training labeled images to the number of spectral features is small. This is because labeled data is expensive, and it is difficult to collect a large number of training samples. For such scenarios, Bandos et al.\  \cite{hyper_RLDA} used regularized LDA in the context of hyperspectral image classification. Du and Nekovel \cite{hyper_CLDA} proposed a Constrained LDA for efficient real-time hyperspectral image classification.

Finally, \textbf{Neighborhood Component Analysis (NCA)} was introduced by Goldberger et al.\  \cite{NCA_NIPS}. Using a linear transform $\mathbf{A}$, NCA aims to find a feature space such that the average leave-one-out k-Nearest Neighbor (k-NN) score in the transformed space is maximized. It can be represented as:
\begin{equation}
\label{eq:NCA}
\mathbf{Z}=\mathbf{A}\mathbf{X}.
\end{equation}

NCA aims to reduce the input dimensionality $N$ by learning the transform $\mathbf{A}$ from the data-set with the help of a differentiable cost function for $\mathbf{A}$ \cite{NCA_NIPS}. However, this cost function is non-convex in nature, and thus the solution obtained may be sub-optimal. 

The transform $\mathbf{A}$ is estimated using a stochastic neighbor selection rule. Unlike the conventional k-NN classifier that estimates the labels using a majority voting of the nearest neighbors, NCA randomly selects neighbors and calculates the expected vote for each class. This stochastic neighbor selection rule is applied as follows. Each point $i$ selects another point as its neighbor $j$ with the following probability:
\begin{equation}
\label{eq:NCA2}
p_{ij}=\frac{e^{-d_{ij}}}{\sum_{k \neq i}^{} e^{-d_{ik}}},
\end{equation}
where $d_{ij}$ is the distance between points $i$ and $j$, and $p_{ii}=0$. 

NCA is used in remote sensing for the classification of hyper-spectral images.  Weizman and Goldberger \cite{NCA_hyper} have demonstrated the superior performance of NCA in the context of images obtained from an airborne visible/infrared imaging spectroradiometer.

\begin{figure}[htb]
\centering
   \subfloat[]{\includegraphics[width=0.09\textwidth]{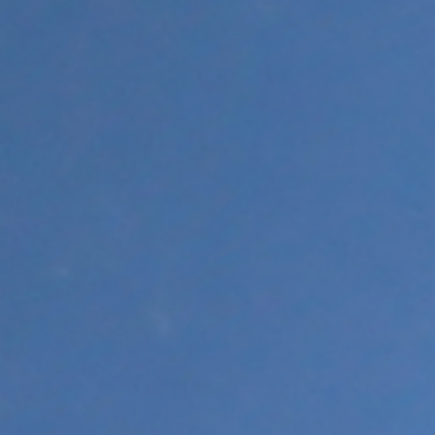}}\hspace{0.01cm}	
   \subfloat[]{\includegraphics[width=0.09\textwidth]{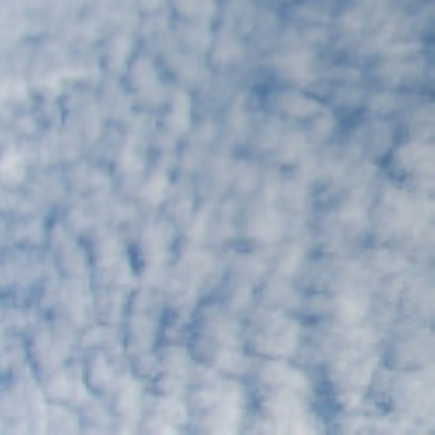}}\hspace{0.01cm}
   \subfloat[]{\includegraphics[width=0.09\textwidth]{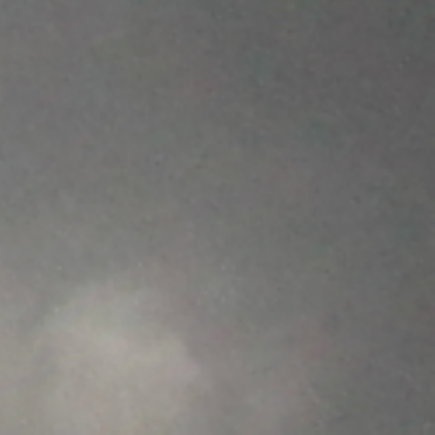}}\hspace{0.01cm}
   \subfloat[]{\includegraphics[width=0.09\textwidth]{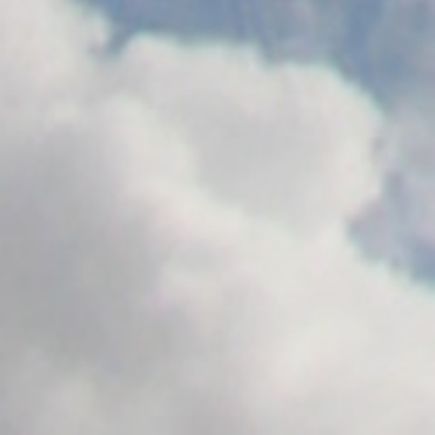}}\hspace{0.01cm}
   \subfloat[]{\includegraphics[width=0.09\textwidth]{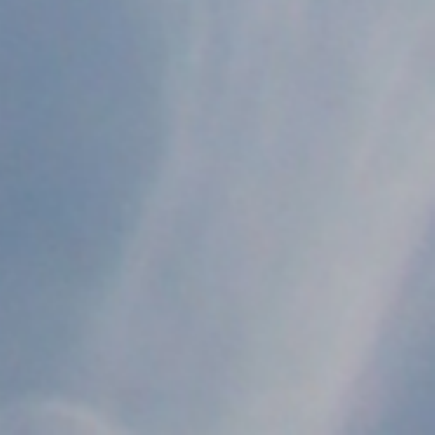}}
\caption{Categories for sky/cloud image patches in SWIMCAT: (a) clear sky, (b) patterned clouds, (c) thick dark clouds, (d) thick white clouds, (e) veil clouds.}
\label{fig:SWIMCAT-images}
\end{figure}

We now illustrate the effect of different DR techniques in the context of ground-based cloud classification. For this purpose, we use the recently released cloud categorization database called SWIMCAT (Singapore Whole-sky IMaging CATegories database)~\cite{ICIP2015b}. Cloud types are properly documented by the World Meteorological Organization (WMO) \cite{WMO_guide}.  The SWIMCAT database\footnote{~SWIMCAT can be downloaded from  \url{http://vintage.winklerbros.net/swimcat.html}} consists of a total of $784$ sky/cloud image patches divided into $5$ visually distinctive categories: clear sky, patterned clouds, thick dark clouds, thick white clouds, and veil clouds. Sample images from each category are shown in Fig.~\ref{fig:SWIMCAT-images}.

\begin{figure*}[tb]
\centering
\setlength\fboxsep{0pt}
\subfloat[PCA]{\includegraphics[height=0.21\textwidth]{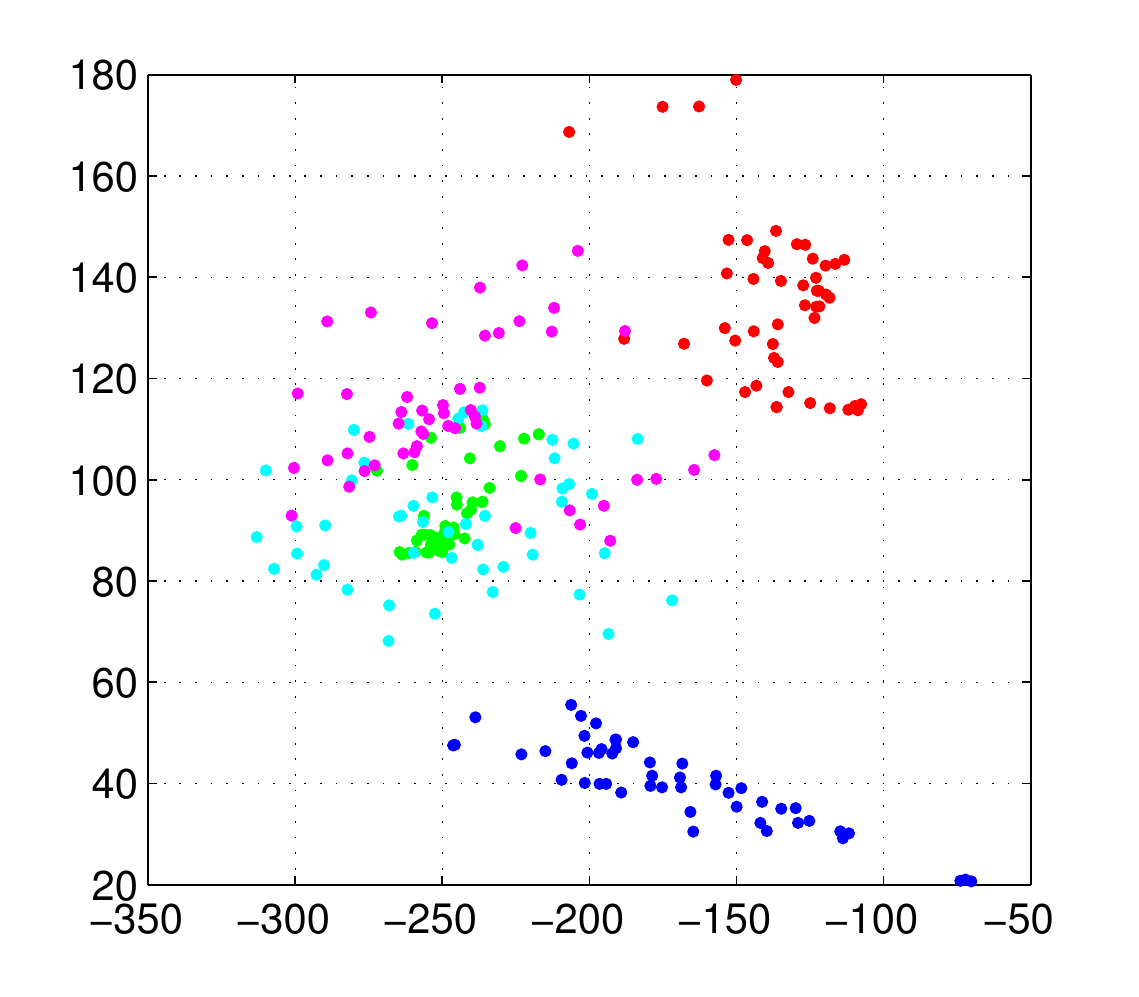}}
\subfloat[FA]{\includegraphics[height=0.21\textwidth]{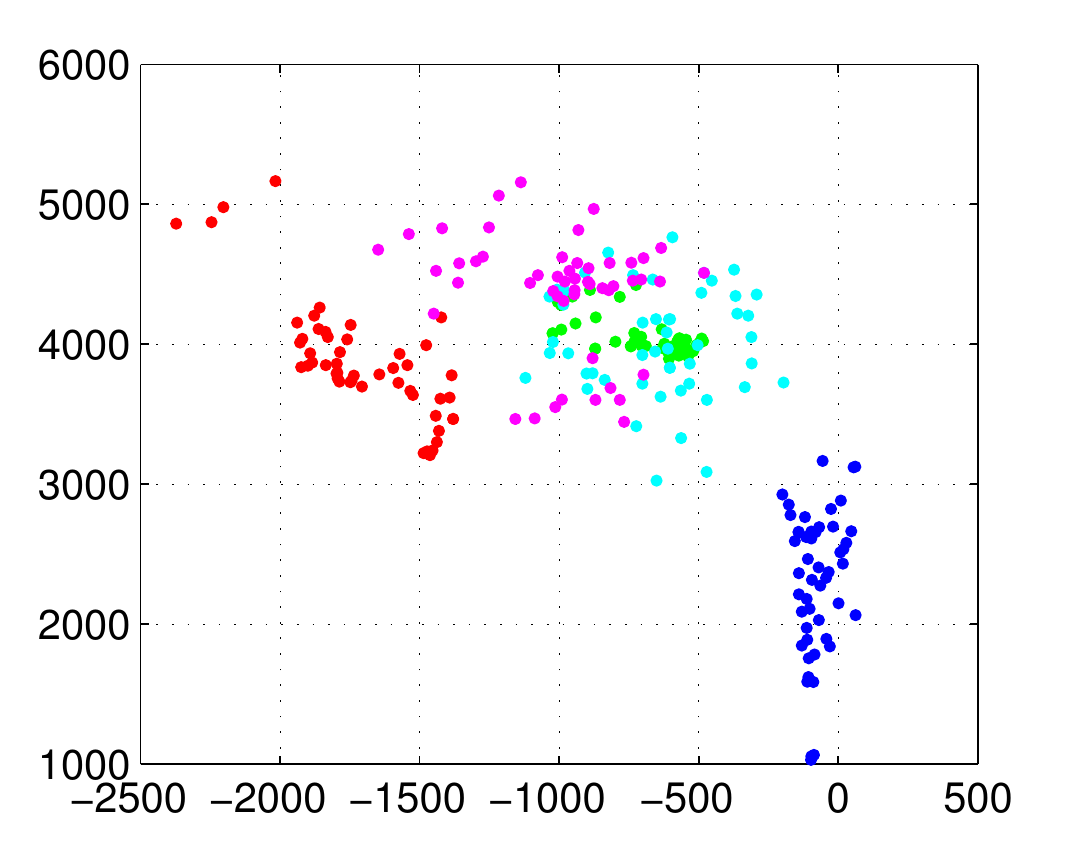}}
\subfloat[LDA]{\includegraphics[height=0.21\textwidth]{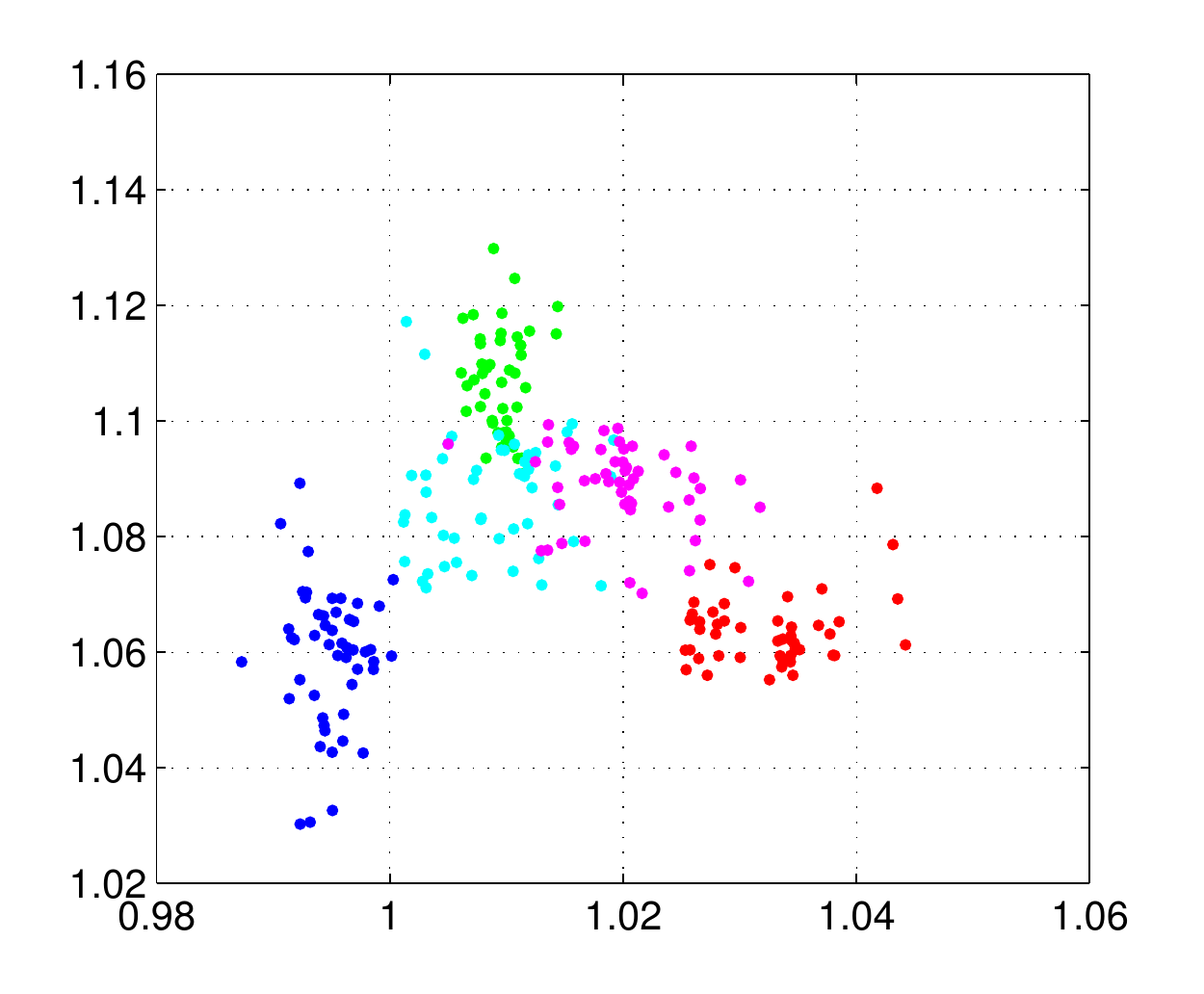}}
\subfloat[NCA]{\includegraphics[height=0.21\textwidth]{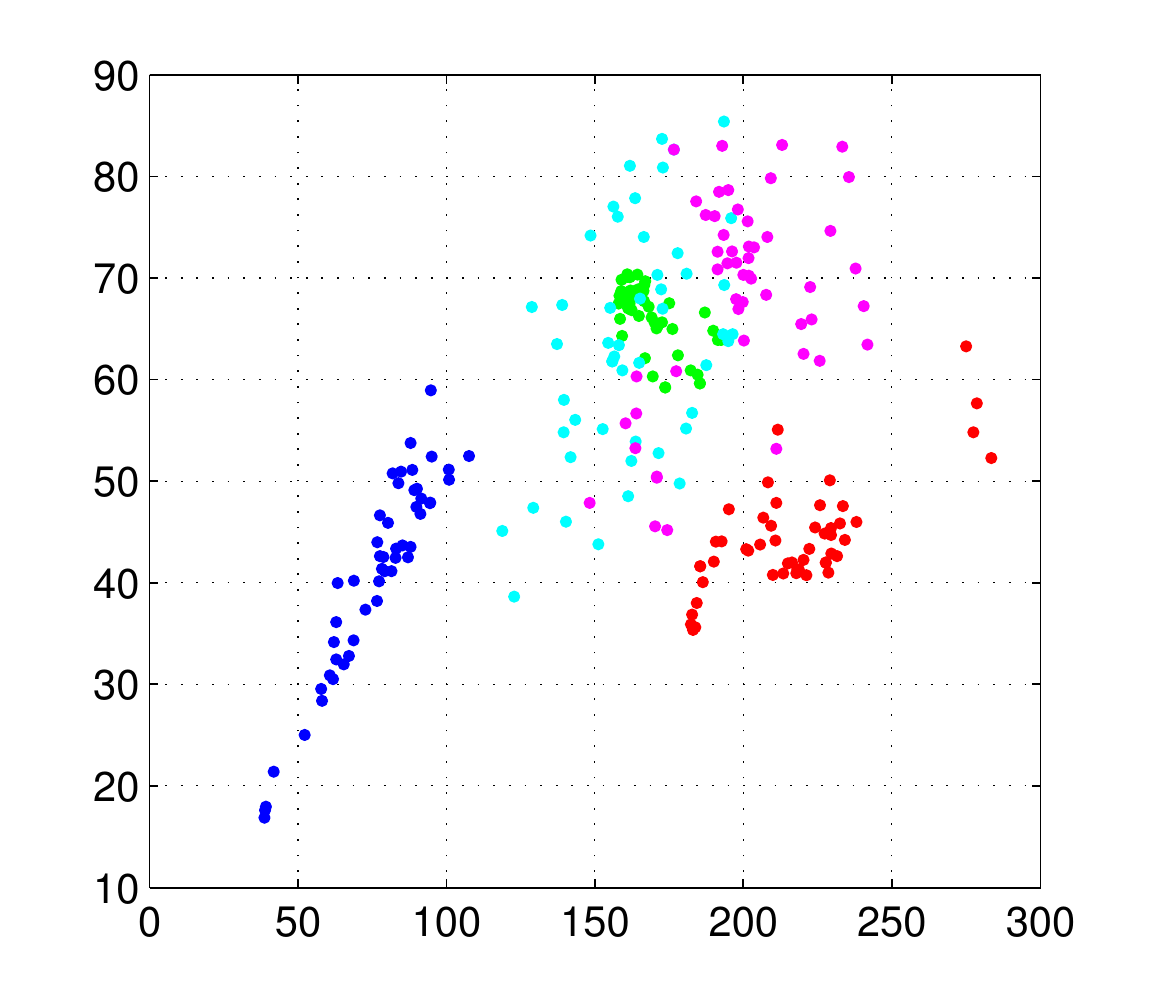}}\\
\caption{Visualization of results from applying four different dimensionality reduction techniques on the SWIMCAT dataset~\cite{ICIP2015b}. The data are reduced from their original 12-dimensional feature space to 2 dimensions in the projected feature space, for a five-class cloud classification problem. The different colors indicate individual cloud classes (red: clear sky; green: patterned clouds; blue: thick dark clouds; cyan: thick white clouds; magenta: veil clouds).}\label{fig:features-plot}
\end{figure*}

We extract the $12$-dimensional Heinle feature (cf.\ Section \ref{sec:RS-feat}) for each image. We randomly select $50$ images from each of the $5$ cloud categories. For easier computation, images are downsampled to a resolution of $32 \times 32$ pixels using bicubic interpolation. Once the feature vectors are generated, the above-mentioned linear DR techniques viz.\ PCA, FA, LDA, NCA are applied on the entire input feature space.

Figure~\ref{fig:features-plot} visualizes the results obtained with the different techniques. The original high-dimensional feature vector is projected onto the primary two principal axes. The different cloud categories are denoted with different colors. We observe that PCA essentially separates the various cloud categories, but veil clouds are scattered in a random manner. PCA and FA are often confused with one another, as they attempt to express the input variables in terms of latent variables. However, we should note that they are distinct methods based on different underlying philosophies, which is exemplified by the results shown in Fig.~\ref{fig:features-plot}. The separation of features in LDA is relatively good as compared to PCA and FA. This is because LDA aims to increase class-separability in addition to capturing the maximum variance. NCA also separates the different classes quite well. In order to further quantify this separability of different classes in the transformed domain, we will present a quantitative analysis in Section~\ref{sec:im-class} below.

\subsection{Sparse Representation Features}
Features based on sparse representation have been widely studied and used in signal processing and computer vision. Different from DR, which provides effective representation in a lower-dimensional subspace, adaptive sparse representation learns a union of subspaces for the data. Compared to fixed sparse models such as the discrete cosine transform (DCT) or wavelets, adaptively learned sparse representation provides improved sparsity, and usually serves as a better discriminator in various tasks including face recognition \cite{yang2007feature}, image segmentation \cite{mairal2008discriminative}, object classification \cite{ramirez2010classification}, and denoising \cite{elad2, wensabres}. Learning-based sparse representation also demonstrates advantages in remote sensing problems such as image fusion \cite{li2013remote} and hyperspectral image classification \cite{chen2011hyperspectral}. 

Several models for sparsity have been proposed in recent years. The most popular one is the \textbf{synthesis model} \cite{elad2}, which suggests that a set of data $\mathbf{X}$ can be modeled by a common matrix $\mathbf{D} \in \mathbb{R}^{N \times K}$ and their respective sparse codes $\mathbf{Z}$:
\begin{equation}
\label{eq:synthesis}
\mathbf{X} = \mathbf{D}\mathbf{Z},\,\,s.t.\,\left \| z_{i} \right \|_{0}\leq s \ll K\;\; \forall \, i,
\end{equation}
where $\left \| . \right \|_{0}$ counts the number of non-zeros, which is upper-bounded by the sparsity level $s$. The codes $\left \{ z_{i} \right \}_{i=1}^{n}$ are sparse, meaning that the maximum number of non-zeros $s$ is much smaller than the code dimensionality $K$. The matrix $\mathbf{D} = \begin{bmatrix}d_{1} \mid d_{2}\mid &...& \mid d_{K}\end{bmatrix}$ is the \emph{synthesis dictionary}, with each $d_{j}$ called an \emph{atom}. This formulation implies that each $x_{i}$ can be decomposed as a linear combination of only $s$ atoms. For a particular $x_i$, the selected $s$ atoms also form its basis. In other words, data that satisfies such a sparse model lives in a union of subspaces spanned by only a small number of selected atoms of $\mathbf{D}$ due to sparsity. The generalized synthesis model  allows for small modeling errors in the data space, which is normally more practical \cite{yang2007feature, elad2}.

Given data $\mathbf{X}$, finding the ``optimal'' dictionary is well-known as the synthesis dictionary learning problem. Since the problem is normally non-convex, and finding the exact solution is NP-hard, various approximate methods have been proposed and have demonstrated good empirical performance. Among those, the K-SVD algorithm \cite{elad2} has become very popular due to its simplicity and efficiency. For a given $\mathbf{X}$, the K-SVD algorithm seeks to solve the following optimization problem:
\begin{equation}
\label{eq:ksvd}
\min_{\mathbf{D},\mathbf{Z}}\: \left \| \mathbf{X} - \mathbf{D}\mathbf{Z} \right \|_{F}^{2}\; \: s.t.\; \:  \left \| z _{i} \right \|_{0}\leq s\; \: \forall \,  i,\;\left \| d _{j} \right \|_{2} = 1\;\forall \, j,
\end{equation}
where $\left \| \mathbf{X} - \mathbf{D}\mathbf{Z} \right \|_{F}^{2}$ represents the modeling error in the original data domain. To solve this joint minimization problem, the algorithm alternates between sparse coding (solving for $\mathbf{Z}$, with fixed $\mathbf{D}$) and dictionary update (solving for $\mathbf{D}$, with fixed $\mathbf{Z}$) steps. K-SVD adopts Orthogonal Matching Pursuit (OMP) \cite{omp} for sparse coding and updates the dictionary atoms sequentially, while fixing the support of corresponding $\mathbf{Z}$ component by using Singular Value Decomposition (SVD).

Besides synthesis dictionary learning, there are learning algorithms associated with other models, such as \textbf{transform learning} \cite{sabres}. Different from synthesis dictionary learning, which is normally sensitive to initialization, the transform learning scheme generalizes the use of conventional analytical transforms such as DCT or wavelets to a regularized adaptive transform $\mathbf{W}$ as follows:
\begin{equation}
\label{eq:transform}
\min_{\mathbf{W},\mathbf{Z}}\: \left \| \mathbf{W}\mathbf{X} - \mathbf{Z} \right \|_{F}^{2}+\nu(\mathbf{W})\; \: s.t.\; \:  \left \| z_{i} \right \|_{0}\leq s\; \: \forall \,  i,
\end{equation}
where $\left \| \mathbf{W} \mathbf{X} - \mathbf{Z} \right \|_{F}^{2}$ denotes the modeling error in the adaptive transform domain. Function $\nu \left ( . \right )$ is the regularizer for $\mathbf{W}$ \cite{sabres}, to prevent trivial and badly-conditioned solutions. The corresponding algorithm \cite{sabres, wensabres} provides exact sparse coding and a closed-form transform update with lower complexity and faster convergence, compared to the popular K-SVD.

In sparse representation, the sparse codes are commonly used as features for various tasks such as image reconstruction and denoising. More sophisticated learning formulations also include the learned models (dictionaries, or transforms) as features for applications such as segmentation and classification. 

\begin{figure}[htb]
\centering
   \subfloat[]{\includegraphics[height=1.3in]{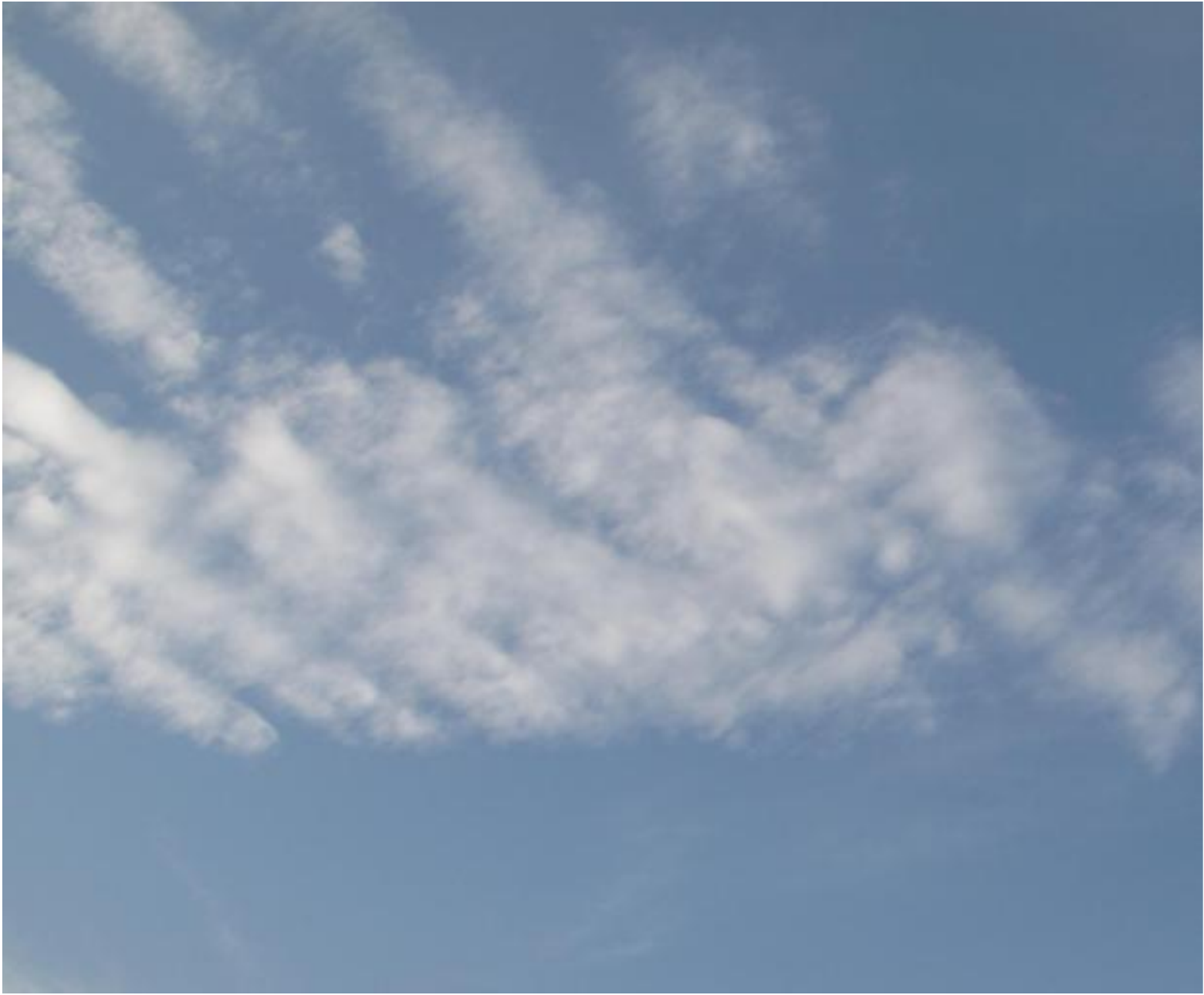}}	
   \subfloat[]{\includegraphics[height=1.3in]{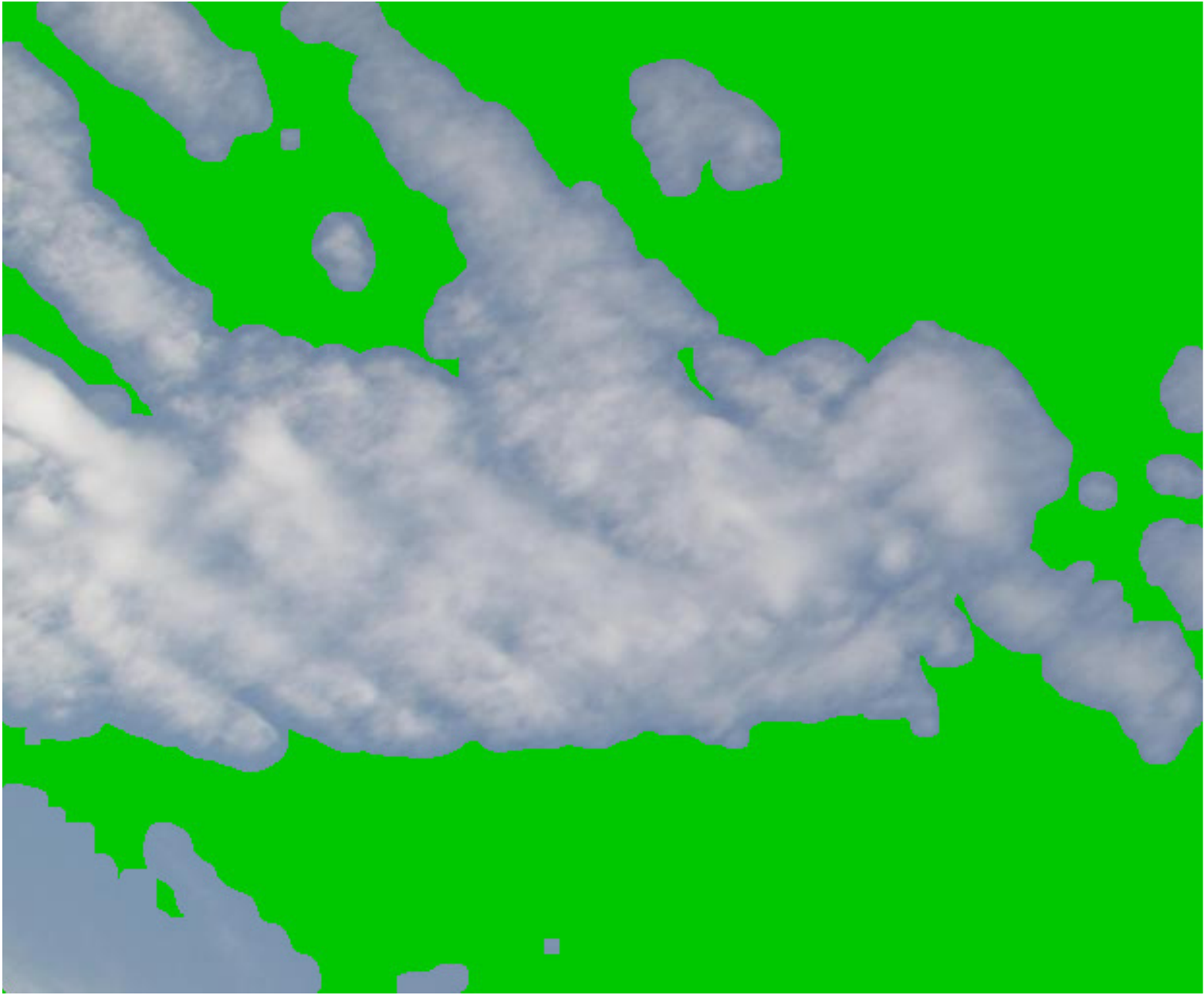}} \hspace{0.02cm}  
   \subfloat[]{\includegraphics[height=1.5in]{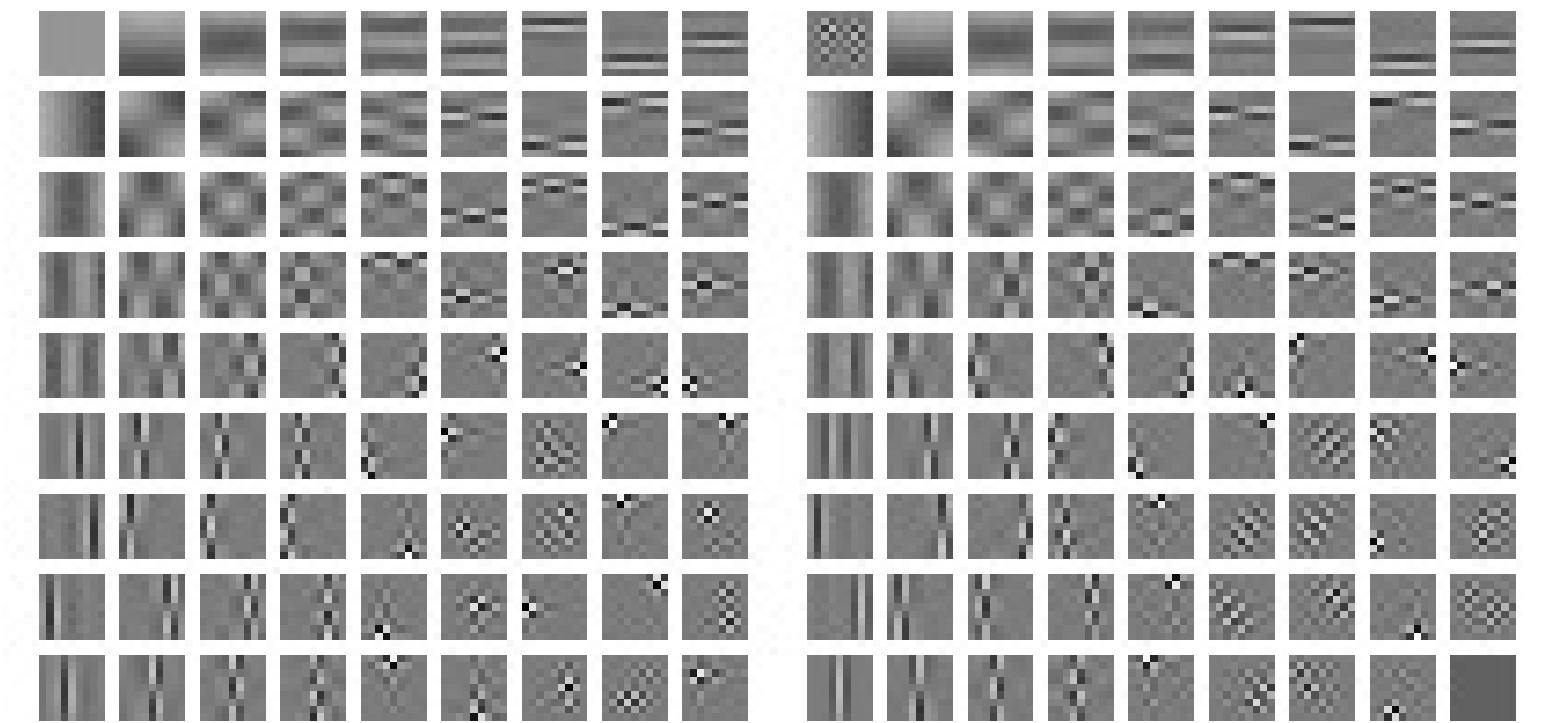}} 
\caption{Cloud and sky segmentation via learning OCTOBOS sparse representation: (a) Original image, (b) input image with original pixels clustered as Cloud, and green pixels clustered as Sky, and (c) learned two-class OCTOBOS, with each row visualized as patches in separate blocks.}
\label{fig:srfeatures}
\end{figure}

Figure~\ref{fig:srfeatures} provides a simple cloud/sky image segmentation example using OCTOBOS \cite{wensabres}, which learns a union of sparsifying transforms, to illustrate and visualize the usefulness of sparse features. We extract $9 \times 9$ overlapping image patches from the ground-based sky image shown in Fig.~\ref{fig:srfeatures}(a). The color patches are converted to gray-scale and vectorized to form the 81-dimensional data vectors. 
The OCTOBOS algorithm simultaneously learns a union of two transforms, generates the sparse codes, and clusters the image patches into two classes (i.e., sky class and cloud class) by comparing the modelling errors \cite{wenICIP2014}.
Since the overlapping patches are used, each pixel in the image typically belongs to multiple extracted patches. We cluster a pixel into a particular class by majority voting. The image segmentation result, with pixels belonging to the sky class, is visualized in Fig.~\ref{fig:srfeatures}(b).
In the learning stage, we restrict the sparsity of each vector to be at most $10$ out of $81$. The distinct sparsifiers, or rows of learned OCTOBOS, are visualized as $9 \times 9$ patches in blocks in Fig.~\ref{fig:srfeatures}(c). Both the sparse codes and the learned transform blocks are used as features for clustering in this example. Note that we did not use any other remote-sensing features on top of the OCTOBOS clustering scheme \cite{wenICIP2014}. A hybrid version which combines this with cloud-specific features \cite{ICIP1_2014} may further enhance the segmentation performance.

\section{Applications}
\label{sec4}
In this section, we present applications of the techniques discussed in the previous section for ground-based sky/cloud image analysis and show experimental results. We focus on three main applications: segmentation, classification, and denoising. We show that data-driven machine learning techniques generally outperform  conventional heuristic approaches.

\subsection{Image Segmentation}
Image segmentation refers to the task of dividing an image into several segments, in an attempt to identify  different objects in the image. The problem of image segmentation has been extensively studied in remote sensing for several decades. In the context of ground-based image analysis, image segmentation refers to the segmentation of sky/cloud images obtained by  sky cameras. Cloud segmentation is challenging because of the clouds' non-rigid structure and the high degree of variability in sky illumination conditions. In this section, we will provide illustrative examples of several sky/cloud image segmentation methodologies. 

\begin{figure}[htb]
\centering
   \subfloat[]{\includegraphics[height=0.88in]{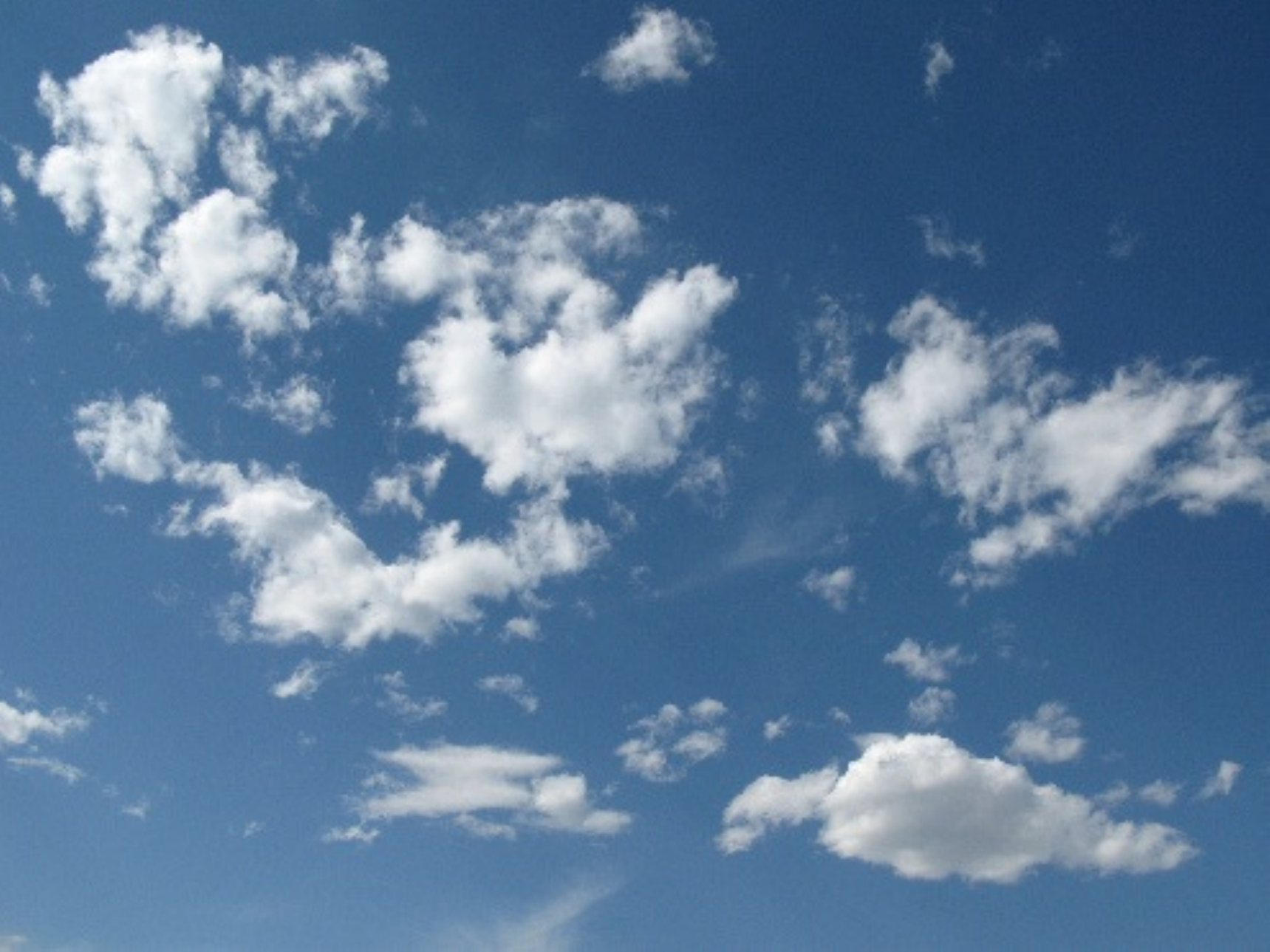}} 	
   \subfloat[]{\includegraphics[height=0.9in]{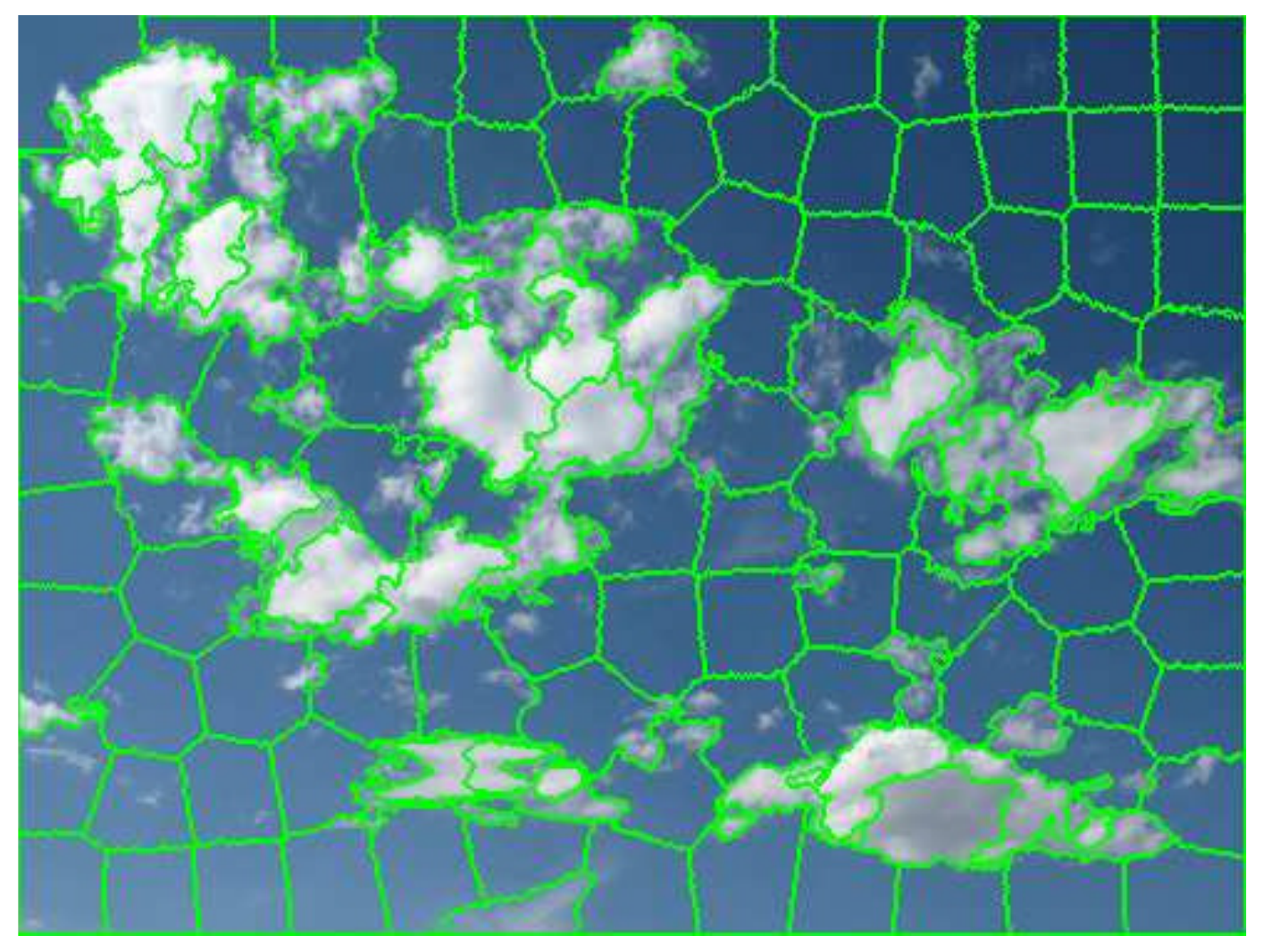}} 
   \subfloat[]{\includegraphics[height=0.88in]{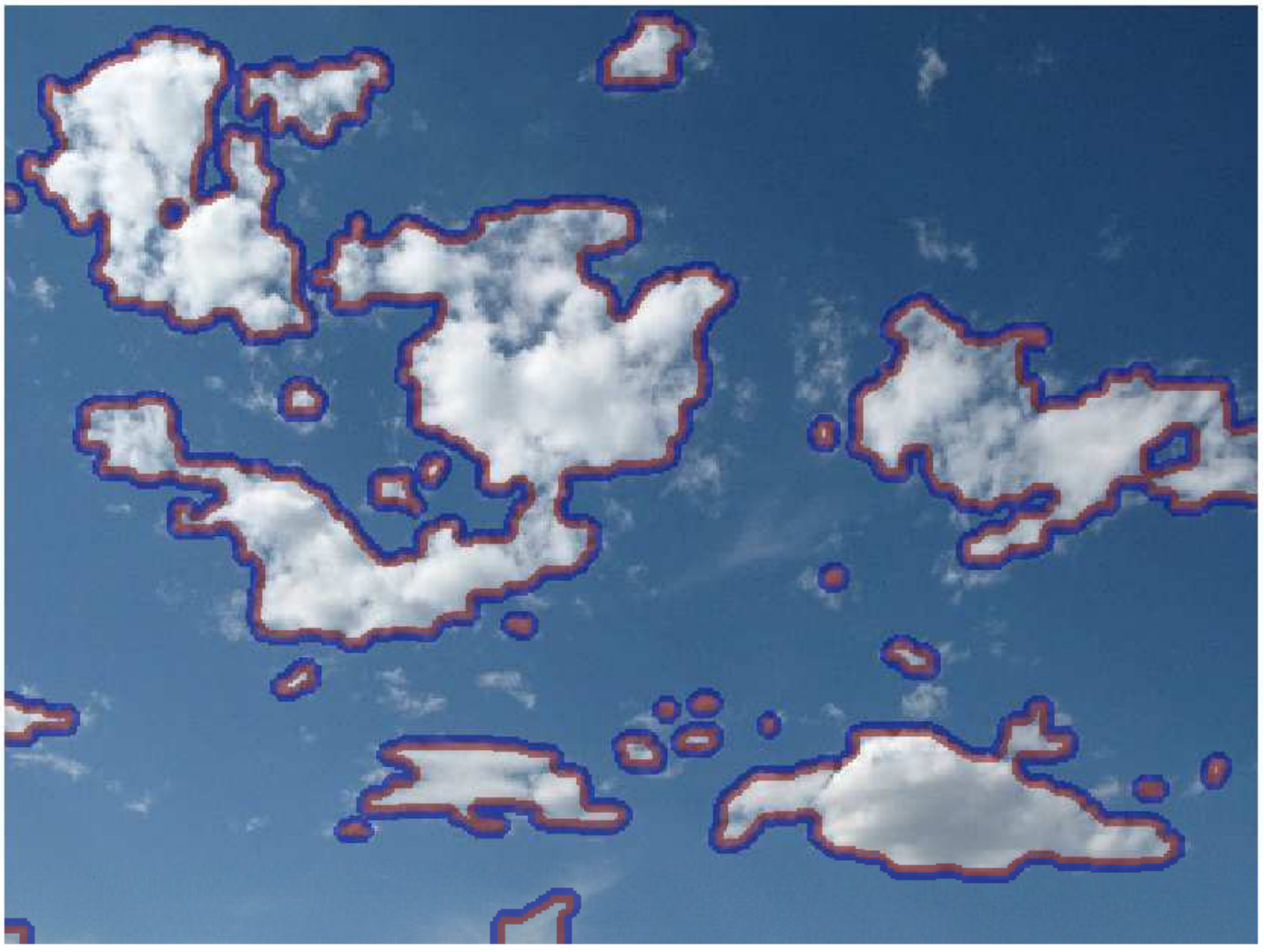}} 
\caption{Illustration of sky/cloud image segmentation using two methods, superpixels and graph-cut. (a) Sample image from HYTA database. (b) Over-segmented image with superpixels. (c) Image segmented using graph-cut.}
\label{fig:seg-based}
\end{figure}

Liu et al.\  \cite{LiuSP2015} use superpixels to identify local homogeneous regions of sky and cloud. In Fig.~\ref{fig:seg-based}, we illustrate the over-segmented superpixel image of a sky/cloud image from the HYTA database~\cite{Li2011}. The generated superpixels respect the image boundaries quite well, and are consistent based on texture and color of sky and cloud regions, respectively. These local regions can thus be used for subsequent machine learning tasks. 

The final sky/cloud binary image can be obtained by thresholding this over-segmented image using a threshold matrix \cite{LiuSP2015}. In addition to superpixels, graph-cut based techniques \cite{Boykov2004,graphcut_PAMI} have also been explored in ground-based image analysis. Liu et al.\ \cite{Liu_AGC} proposed an automatic graph-cut technique in identifying sky/cloud regions. As an illustration, we show the two-level segmented output using automatic graph cut in Fig.~\ref{fig:seg-based}(c).

\begin{figure}[htb]
\centering
   \subfloat[Binary]{\includegraphics[height=0.88in]{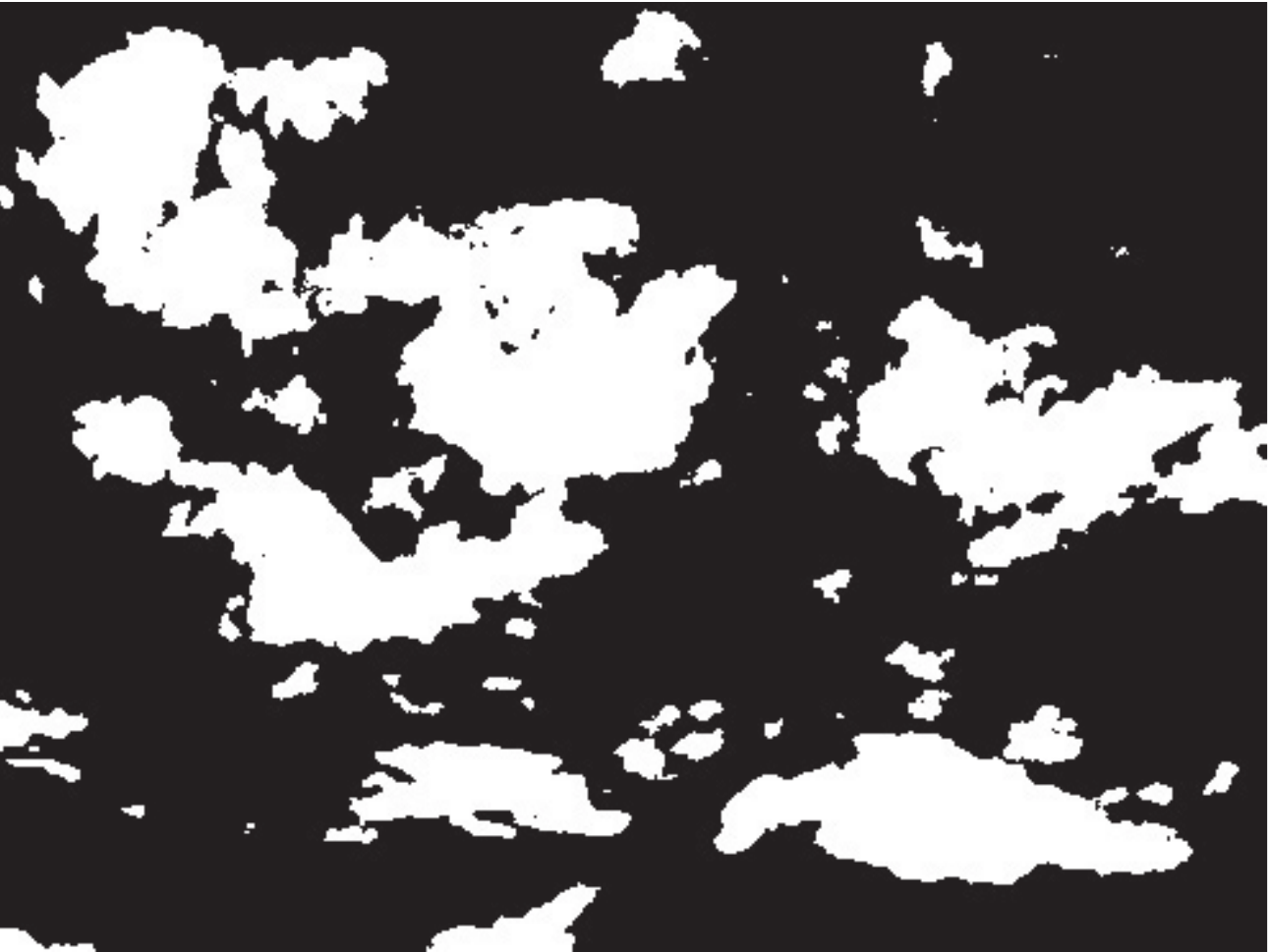}}
   \subfloat[3-level]{\includegraphics[height=0.88in]{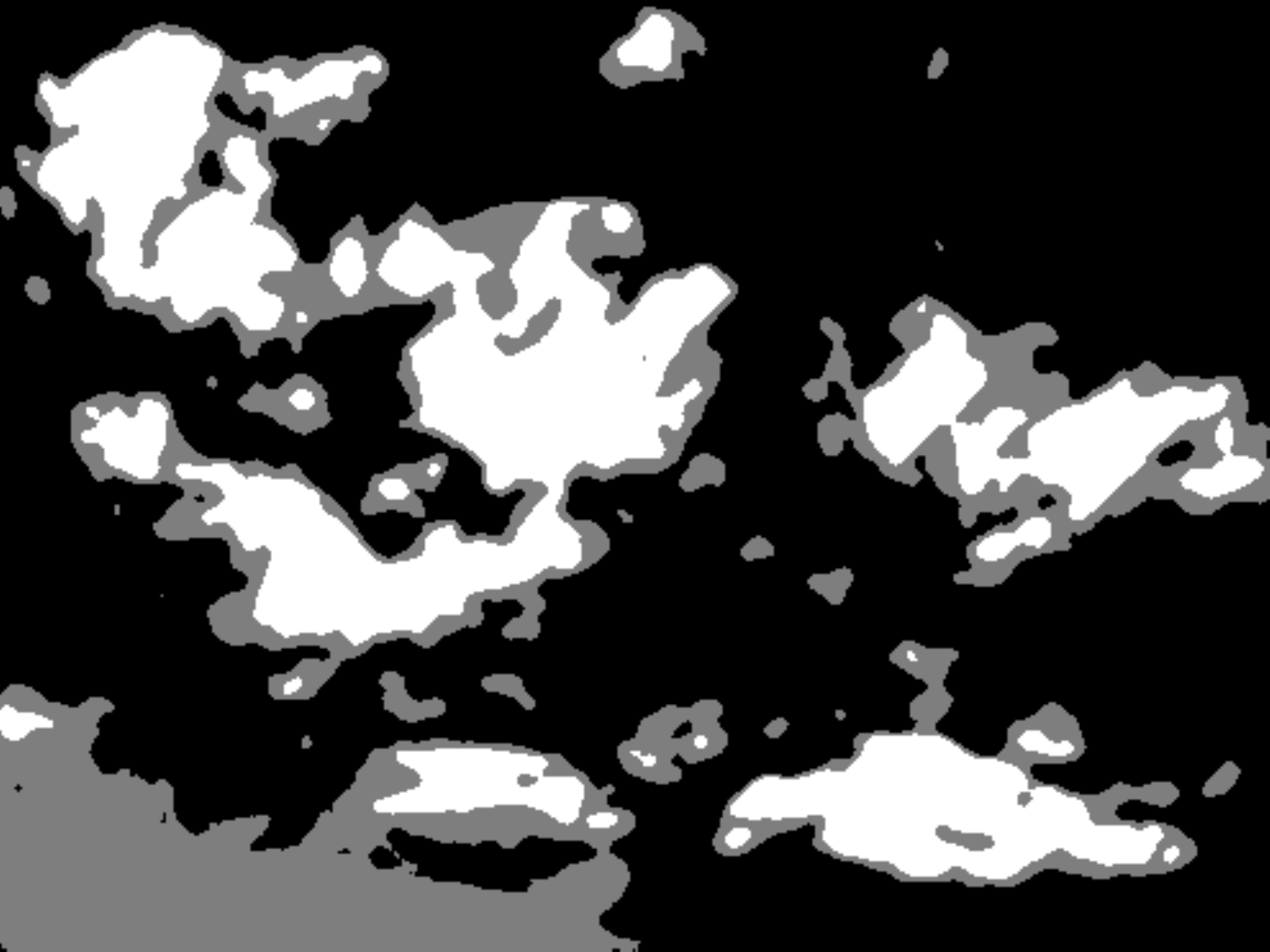}}
   \subfloat[Probabilistic]{\includegraphics[height=0.88in]{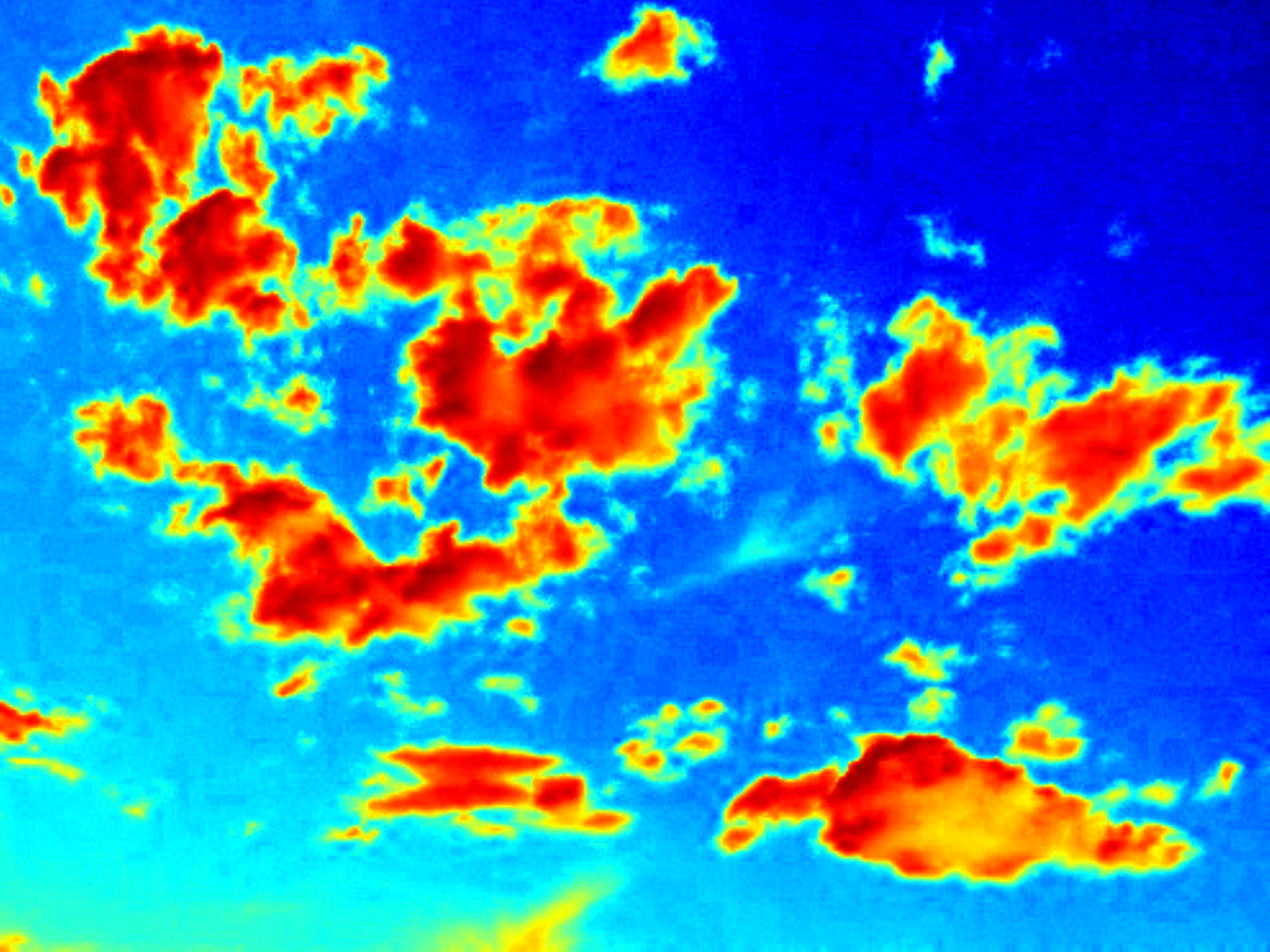}}
\caption{Illustration of sky/cloud image segmentation. (a) Binary (or 2-level) segmentation of a sample input image from HYTA database; (b) 3-level semantic segmentation of sky/cloud image~\cite{ICIP2015a}; (c) probabilistic segmentation of sky/cloud image  \cite{ICIP1_2014}.}\label{fig:UNSUP}
\end{figure}

As clouds do not have any specific shape, and cloud boundaries are ill-defined, several approaches have been proposed that use color as a discriminatory feature. The segmentation can be binary~\cite{Li2011,Souza}, multi-level~\cite{ICIP2015a}, or probabilistic~\cite{ICIP1_2014}. As an illustration, we show these three cases for a sample image of HYTA dataset. Figure~\ref{fig:UNSUP}(a) shows the binary segmentation of a sample input image from the HYTA database~\cite{Li2011}. The process involves thresholding the selected color channel. 

In addition to such binary approaches, a multi-level output image can also be generated. Machine learning techniques involving Gaussian discriminant analysis can be used for such purposes. In \cite{ICIP2015a}, a set of labeled training data is used for a-priori learning of the latent distribution of three labels (clear sky, thin clouds, and thick clouds). We illustrate such 3-level semantic labels of the sky/cloud image in Fig.~\ref{fig:UNSUP}(b).

In addition to 2-level and 3-level output images, a probabilistic segmentation approach is exploited in \cite{ICIP1_2014}, wherein each pixel is assigned a confidence value of belonging to the cloud category. This is illustrated in Fig.~\ref{fig:UNSUP}(c). 

\subsection{Image Classification}
\label{sec:im-class}
In the most general sense, classification refers to the task of categorizing the input data into two (or more) classes. We can distinguish between supervised and unsupervised methods.  The latter identify underlying latent structures in the input data space, and thereby make appropriate decisions on the corresponding labels. In other words, unsupervised methods cluster pixels with similar properties (e.g.\ spectral reflectance). Supervised methods on the other hand, rely on a set of annotated training examples. This training data helps the system to learn the distribution of the labeled data in any dimensional feature space. Subsequently, the learned system is used in predicting the labels of unknown data points.

In remote sensing, k-means, Gaussian Mixture Models (GMM) and swarm optimization are the most commonly used unsupervised classification (clustering) techniques. Ari and Aksoy~\cite{Ari2010} used GMM and particle swarm optimization for hyperspectral image classification. Maulik and Saha~\cite{Maulik2009} used a modified differential evolution based fuzzy clustering algorithm for satellite images. Such clustering techniques are also used in ground-based image analysis.

In addition to supervised and unsupervised methods, Semi-Supervised Learning (SSL) methods are widely used in remote sensing~\cite{SSL_remote}. SSL uses both labeled and unlabeled data in its classification framework. It helps in creating a robust learning framework, which learns the latent marginal distribution of the labels. This is useful in remote sensing, as the availability of labeled data is scarce and manual annotation of data is expensive. One such example is hyperspectral image classification \cite{HSI-SSL}. In addition to SSL methods, models involving sparsity and other regularized approaches are also becoming popular. For example, Tuia et al.~\cite{tuia2015tobe} study the use of non-convex regularization in the context of hyperspectral imaging. 

In ground-based image analysis, image classification refers to categorizing sky/cloud types into various kinds, e.g.\ clear sky, patterned clouds, thick dark clouds, thick white clouds and veil clouds (cf.\ Section~\ref{S:DR}). In order to quantify the accuracy of the separation of data in Fig.~\ref{fig:features-plot}, we use several popular clustering techniques in combination with DR techniques. We use two classifiers for evaluation purposes, namely k-Nearest Neighbors (k-NN) and Support Vector Machine (SVM).  k-NN is a non-parametric classifier, wherein the output label is estimated using a majority voting of the labels of a neighborhood. Support Vector Machine (SVM) is a parametric method that generates a hyperplane or a set of hyperplanes in the vector space by maximizing the margin between  classifiers to the nearest neighbor data. 

\begin{figure}[htb]
\centering
   \includegraphics[width=0.5\textwidth]{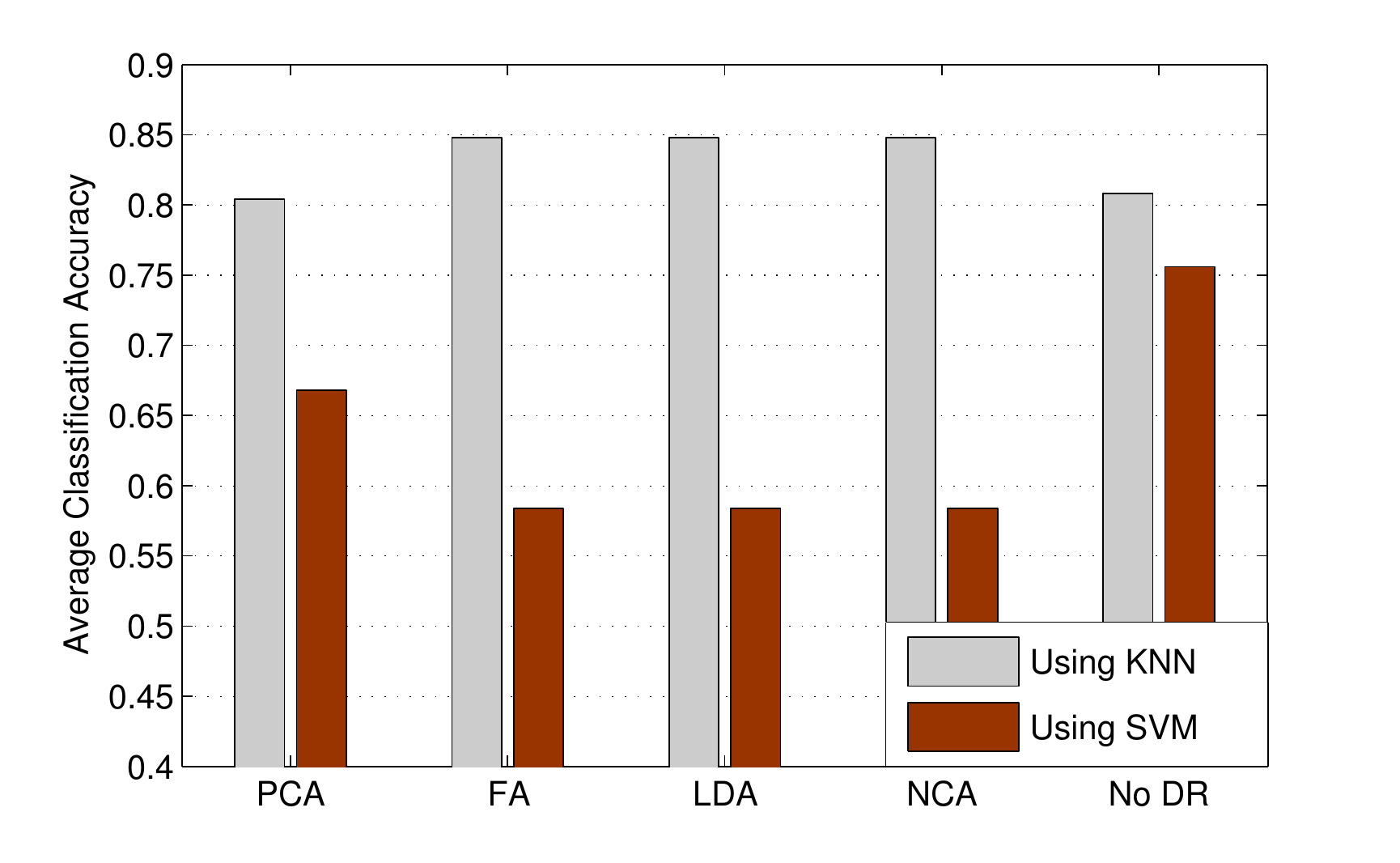}
\caption{Average multi-class classification accuracy using Heinle features for cloud patch categorization for different methods.}\label{fig:classification}
\end{figure}

We evaluate five distinct scenarios: (a) PCA, (b) FA, (c) LDA, (d) NCA, (e) no dimensionality reduction, and report the classification performances of both k-NN and SVM in each of these cases. We again use the SWIMCAT~\cite{ICIP2015b} database for evaluation purposes. The training and testing sets consist of random selections of $50$ distinct images. All images are downsampled to $32 \times 32$ pixels for faster computation. Using the $50$ training images for each of the categories, we compute the corresponding projection matrix for PCA, FA, LDA, and NCA. We use the reduced 2-dimensional Heinle feature for training a k-NN/SVM classifier for scenarios (a-d). We use the original 12-dimensional vector for training the classifier model for scenario (e). In the testing stage, we obtain the projected 2-D feature points using the computed projection matrix, followed by a k-NN/SVM classifier for classifying the test images into individual categories. The average classification accuracies across the $5$ classes are shown in Fig.~\ref{fig:classification}.

The k-NN classifier achieves better performance than the SVM classifier in all of the cases. From the 2-D projected feature space (cf.\ Fig.~\ref{fig:features-plot}) it is clear that the data points belonging to an individual category lie close to each other. However, it is difficult to separate the different categories using hyperplanes in 2-D space. We observe that the complexity of the linear SVM classifier is not sufficient to separate the individual classes. k-NN performs relatively better in this example. Amongst the different DR techniques, LDA and NCA work best with the k-NN classifier. This is because these methods also use the class labels to obtain maximum inter-class separability. Moreover, the performance without prior dimensionality reduction performs comparably well. In fact, the SVM classifier provides increasingly better results when the feature space has higher dimensionality. This shows that further applications of DR on top of extracting remote sensing features may not be  necessary in a classification framework. Of course, dimensionality reduction significantly reduces the computational complexity.

\subsection{Adaptive Denoising}
Image and video denoising problems have been heavily studied in the past, with various denoising methods proposed \cite{secretdenoising}. Denoting the true signal (i.e., clean image or video) as $x$, the measurement $y$ is usually corrupted by additive noise $e$ as 
\begin{equation}
\label{noisy}
y = x + e.
\end{equation}
The goal of denoising is to obtain an estimate $\tilde{x}$ from the noisy measurement $y$ such that $\left\| \tilde{x} - x \right\|$ is minimized. Denoising is an ill-posed problem. Thus, certain regularizers, including sparsity, underlying distribution, and self-similarity, are commonly used to obtain the best estimate $\tilde{x}$.

Early approaches of denoising used fixed analytical transforms, simple probabilistic models \cite{richardson1972bayesian}, or neighborhood filtering \cite{lee1981refined}. Recent non-local methods such as BM3D \cite{bm3d} have been shown to achieve excellent performance, by combining some of these conventional approaches. In the field of remote sensing, Liu et al.\ \cite{PDE-denoise} used partial differential equations for denoising multi-spectral and hyper-spectral images. Yu and Chen \cite{GMCA-denoise} introduced Generalized Morphological Component Analysis (GMCA) for denoising satellite images.

Recently, machine learning based denoising methods have received increasing interest. Compared to fixed models, adaptive sparse models \cite{elad2, sabres} or probabilistic models \cite{gmm2} have been shown to be more powerful in image reconstruction. The popular sparsity-based methods, such as K-SVD \cite{elad2} and OCTOBOS \cite{wensabres}, were introduced in Section \ref{sec3}. Besides, adaptive GMM-based denoising \cite{gmm2} also provides promising performance by learning a GMM from the training data as regularizer for denoising, especially in denoising images with complicated underlying structures. 

While these data-driven denoising methods have become popular in recent years, the usefulness of signal model learning has rarely been explored in remote sensing or ground-based image analysis, which normally generates data with certain unique properties. Data-driven methods can potentially be even more powerful for representing such signals than conventional analytical models. 

\begin{figure}[htb]
\centering
   \subfloat[]{\includegraphics[width=0.49\columnwidth]{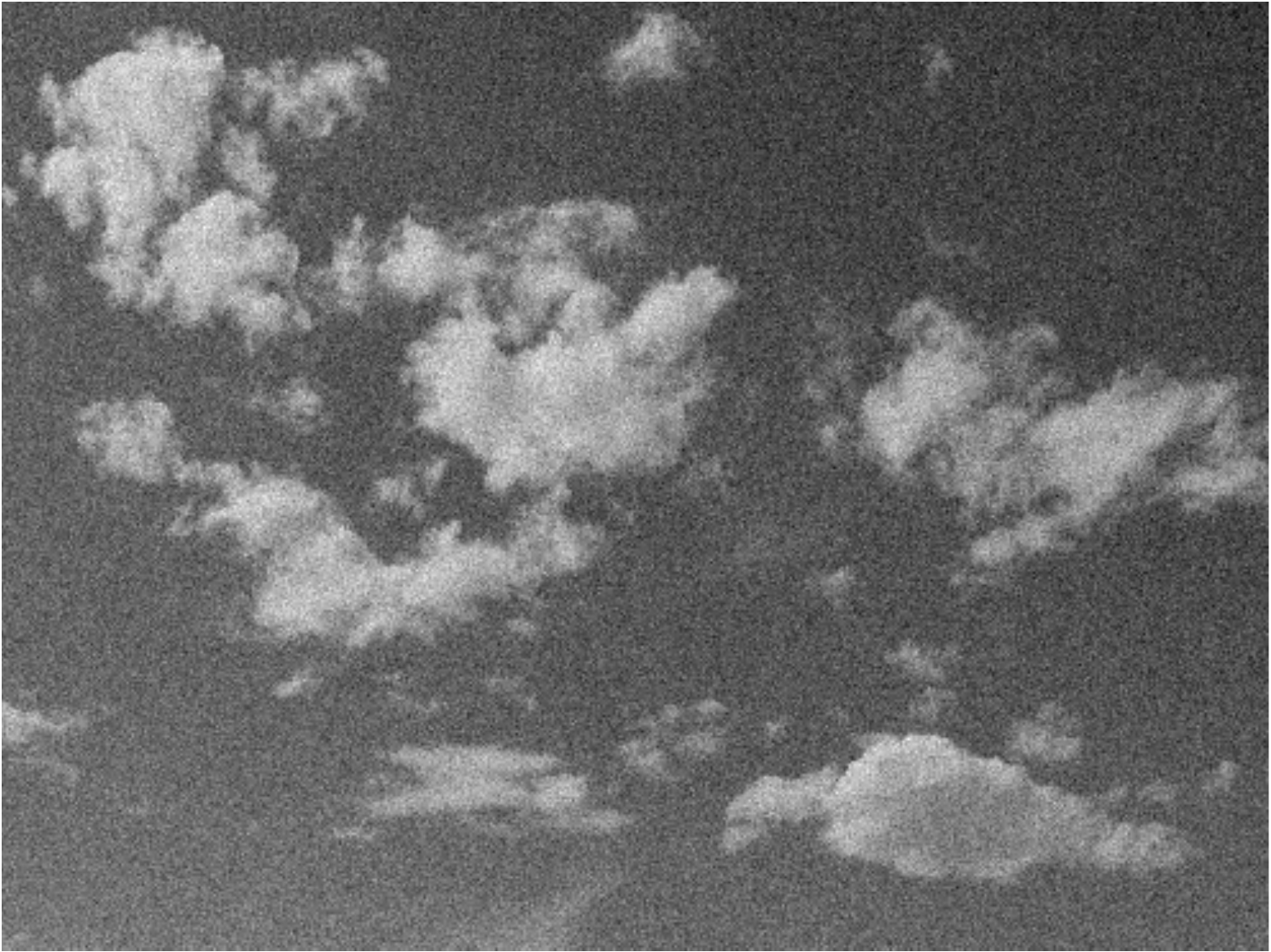}}
   \subfloat[]{\includegraphics[width=0.49\columnwidth]{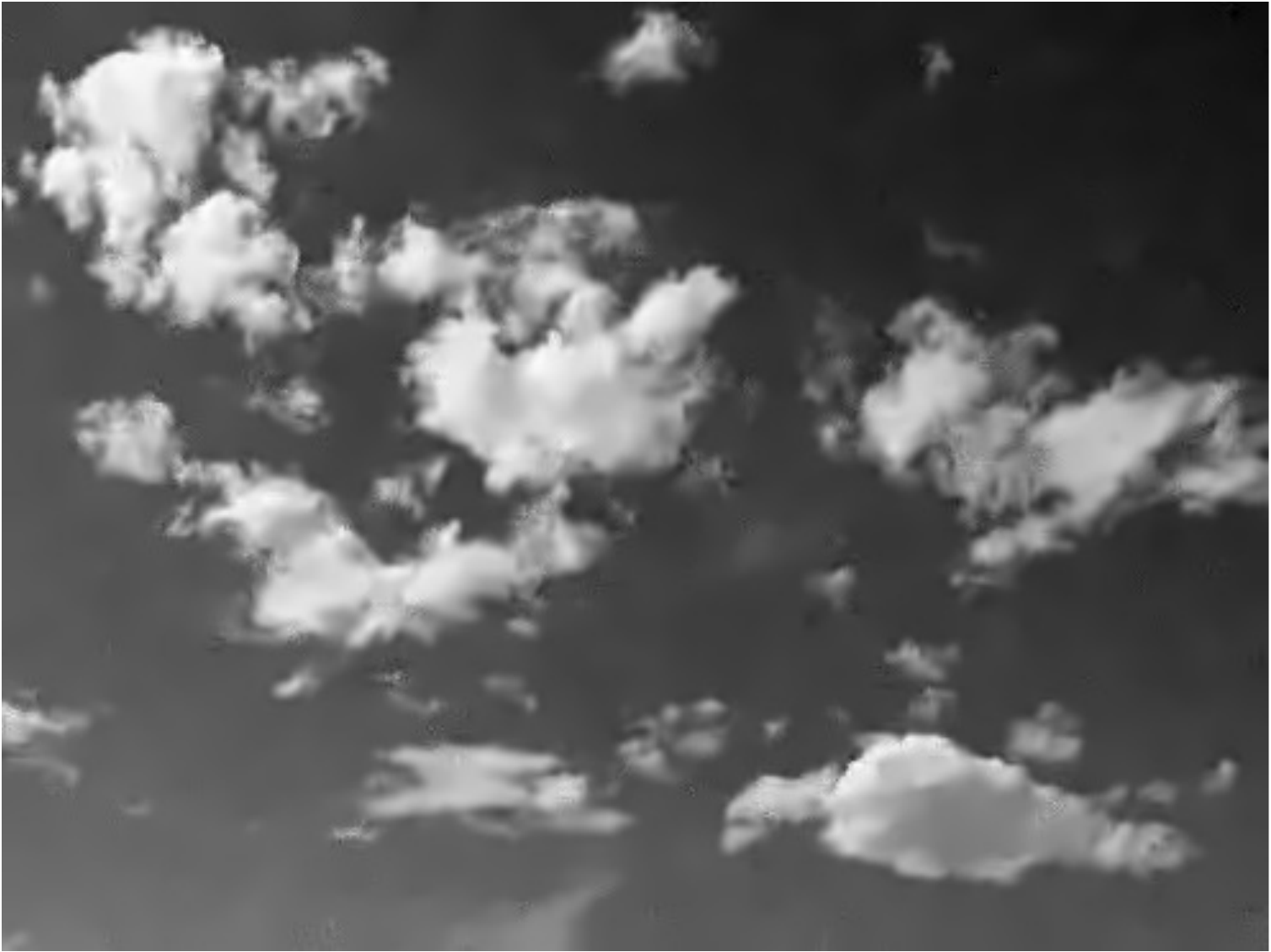}}
\caption{Ground-based image denoising result: (a) Noisy cloud image (PSNR = $22.1$dB), (b) Denoised image (PSNR = $33.5$dB) obtained by using a GMM-based algorithm.}\label{GMMdenoise}
\end{figure}

We now illustrate how various popular learning-based denoising schemes can be applied to ground-based cloud images. The same cloud image from the HYTA database \cite{Li2011} shown in Fig.~\ref{fig:seg-based}(a) is used as an example and serves as ground truth. We synthetically add zero-mean \ Gaussian noise with $\sigma = 20$ to the clean data. The obtained noisy image has a PSNR of $22.1$dB and is shown in Fig.~\ref{GMMdenoise}(a).

\begin{figure}[htb]
\centering
   \includegraphics[width=0.85\columnwidth]{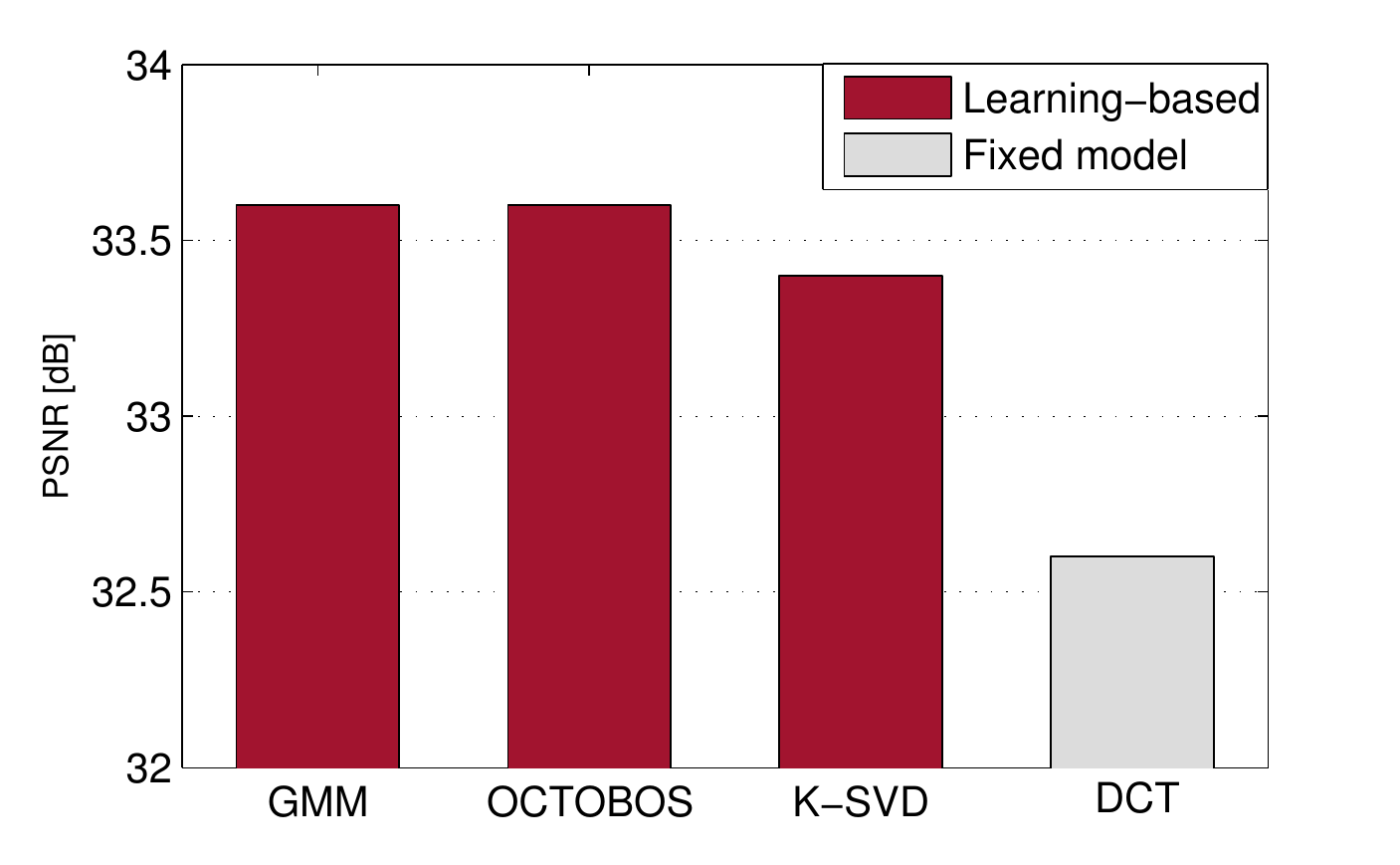}
\caption{PSNR values for denoising with OCTOBOS, GMM, K-SVD, and DCT dictionary.}\label{regulardenoise}
\end{figure}

Figure~\ref{regulardenoise} provides the denoising performance comparison using several popular learning-based denoising schemes, including GMM \cite{gmm2}, OCTOBOS \cite{wensabres}, and K-SVD \cite{elad2}. The quality of the denoised image is measured by Peak Signal-to-Noise Ratio (PSNR) as the objective metric (the clean image has infinite PSNR value). As a comparison, we also include the denoising PSNR by applying a fixed overcomplete DCT dictionary \cite{elad2}. DCT is an analytical transform commonly used in image compression. For a fair comparison, we maintain the same sparse model richness, by using a $256 \times 64$ transform in OCTOBOS, and $64 \times 256$ dictionaries in K-SVD and DCT methods. For GMM, we follow the default settings in the publicly available software \cite{gmm2}. 

As illustrated in Fig.~\ref{regulardenoise}, learning-based denoising methods clearly provide better denoised PSNRs than DCT-based method, with an average improvement of $1.0$ dB. Among all of the learning-based denoising algorithms, K-SVD and OCTOBOS are unsupervised learning methods using image sparsity. OCTOBOS additionally features a clustering procedure in order to learn a structured overcomplete sparse model. GMM is a supervised learning method, which is pre-trained with a standard image corpus. In our experiment, OCTOBOS and GMM perform slightly better than K-SVD, since they are using either a more complicated model or supervised learning. The denoising result using the GMM-based method is shown in Fig.~\ref{GMMdenoise}(b).

WSIs continuously generate large-scale cloud image data which need to be processed efficiently. Although learning-based algorithms can provide promising performance in applications such as denoising, most of them are batch algorithms. Consequently, the storage requirements of batch methods such as K-SVD and OCTOBOS increase with the size of the dataset; besides, processing real-time data in batch mode translates to latency. Thus, online versions of learning-based methods \cite{Mai, sawenbres1} are needed to process high-resolution WSI data. These online learning schemes are more scalable to big-data problems, by taking advantage of stochastic learning techniques. 

\begin{figure}[htb]
\centering
   \subfloat[]{\includegraphics[width=0.49\columnwidth]{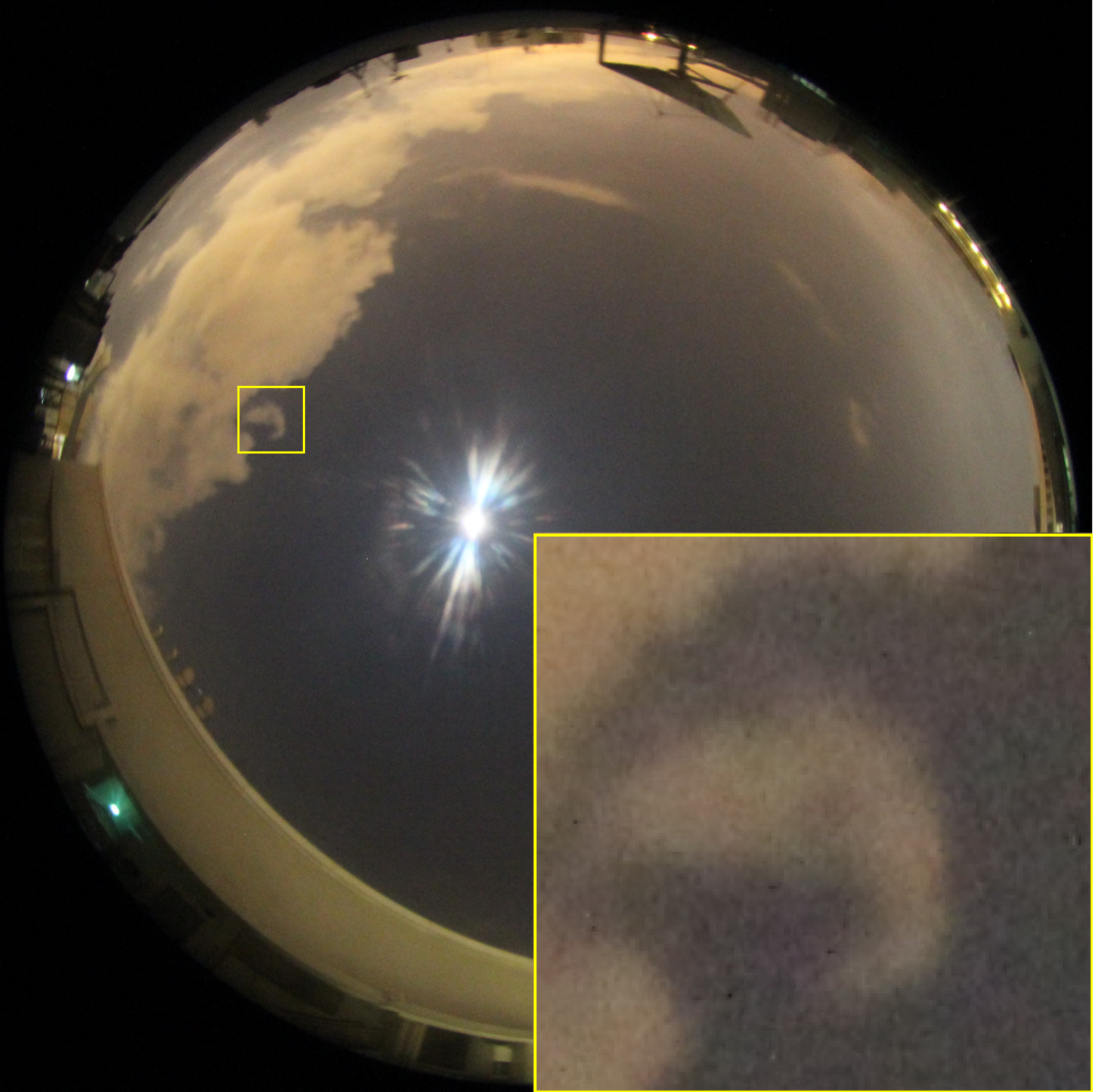}}
   \subfloat[]{\includegraphics[width=0.49\columnwidth]{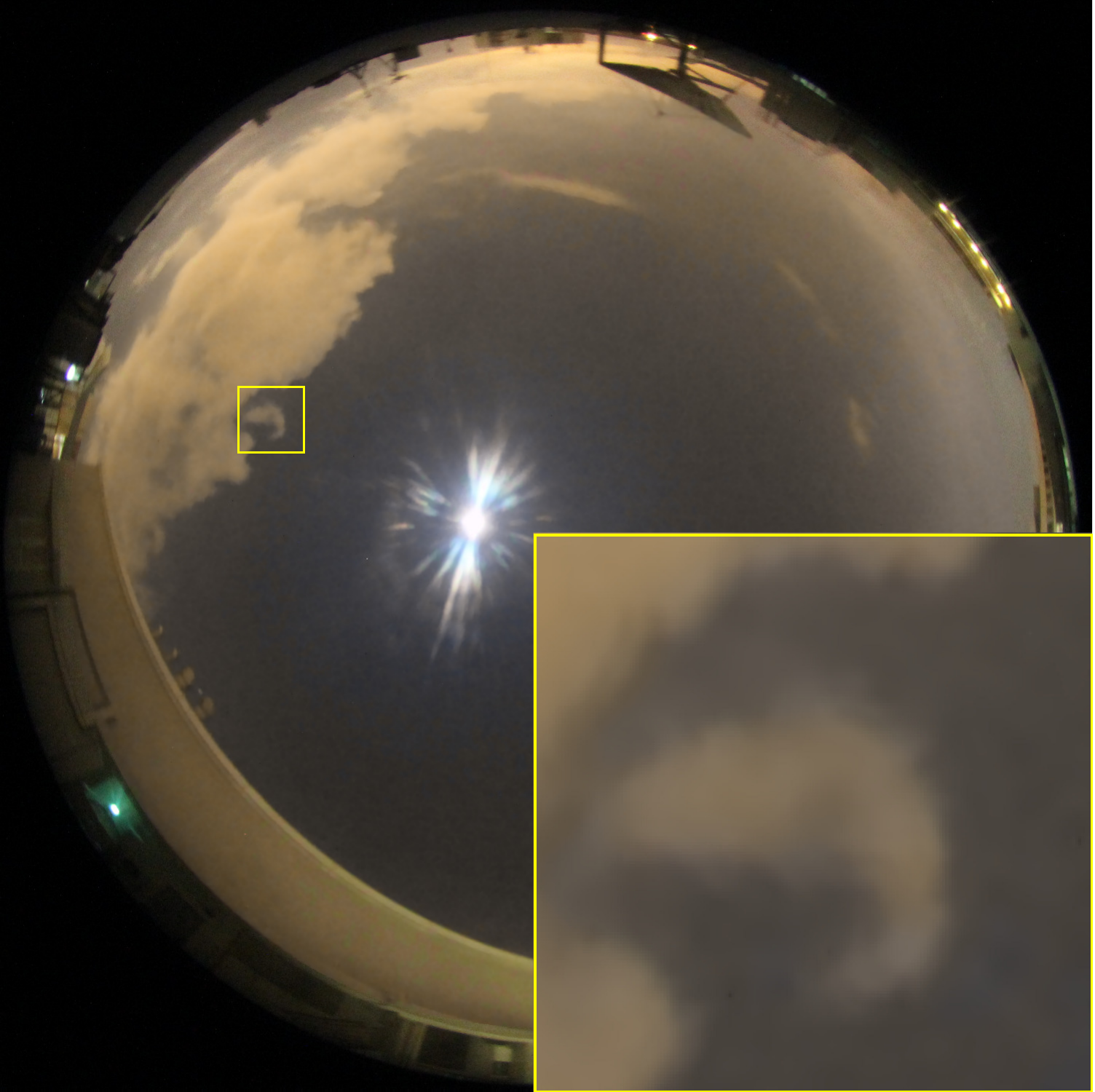}}
\caption{Real large-scale night-time cloud/sky image denoising result, with regional zoom-in for comparison: (a) real noisy cloud image; (b) denoised image obtained using an online transform learning based denoising scheme  \cite{saiGlobal2014}.}\label{OnlineTL_real}
\end{figure}

Here, we show an example of denoising a color image measurement of $3000 \times 3000$ pixels generated by WAHRSIS at night, using online transform learning \cite{saiGlobal2014}. The denoising results are illustrated in Fig.~\ref{OnlineTL_real}. Note that such a method is also capable of processing real-time high-dimensional data \cite{vidolsat}. Thus it can be easily extended to applications involving multi-temporal satellite images and multispectral data in remote sensing.

\section{Conclusion}
In this tutorial paper, we have provided an overview of recent developments in machine learning for remote sensing, using examples from ground-based image analysis. Sensing the earth's atmosphere using high-resolution ground-based sky cameras provides a cheaper, faster, and more localized manner of data acquisition. 

Because of the inherent high-dimensionality of the data, it is expensive to directly use raw data for analysis. We have introduced several feature extraction techniques and demonstrated their properties using illustrative examples. We have also provided extensive experimental results in segmentation, classification, and denoising of sky/cloud images. Several techniques from machine learning and computer vision communities have been adapted to the field of remote sensing and often outperform conventional heuristic approaches.

\bibliographystyle{IEEEbib}


\begin{IEEEbiography}
[{\includegraphics[width=1in,height=1.25in,clip,keepaspectratio]{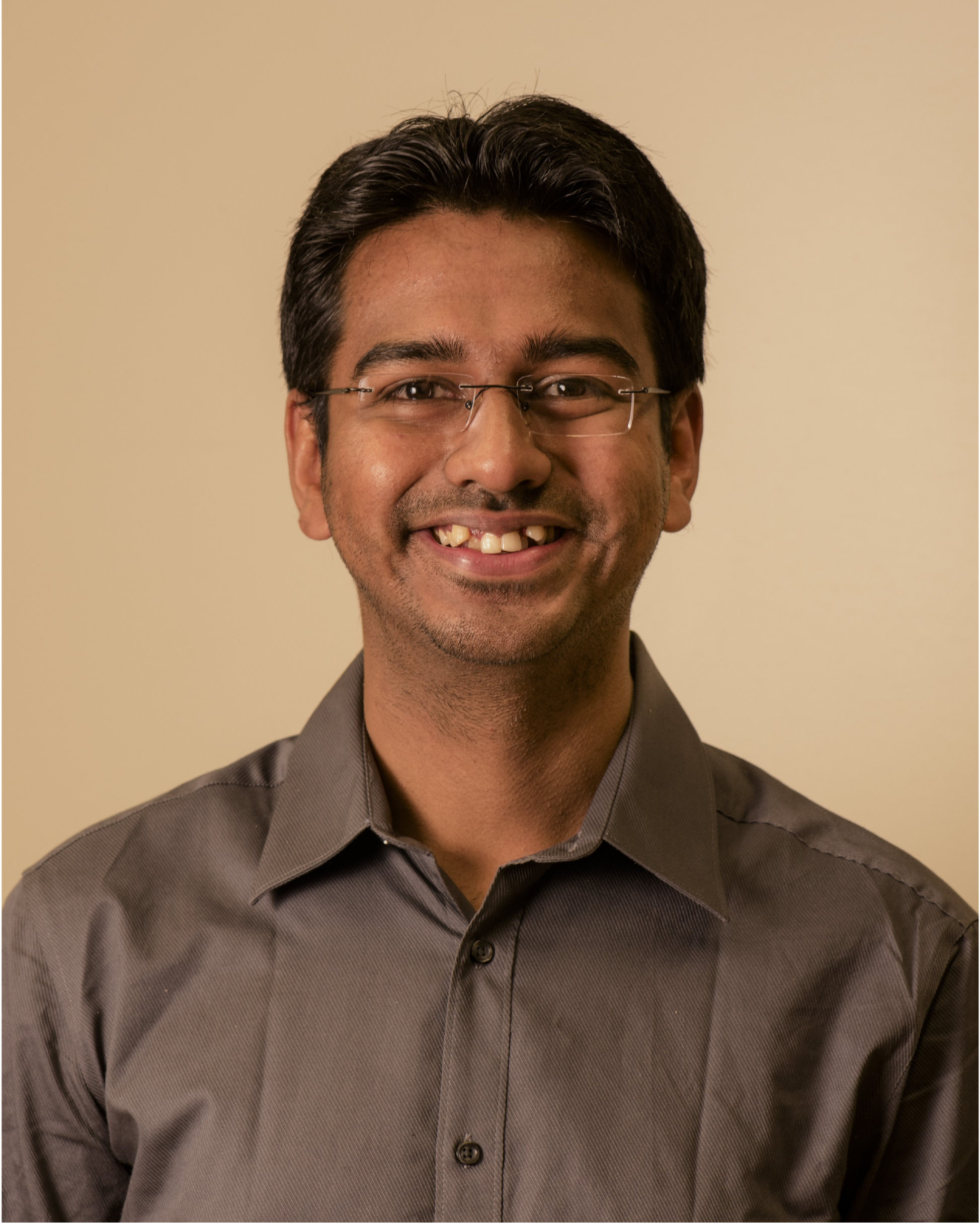}}]{Soumyabrata Dev} (S'09) graduated summa cum laude from National Institute of Technology Silchar, India with a B.Tech.\ in Electronics and Communication Engineering in 2010. Subsequently, he worked in Ericsson as a network engineer from 2010 to 2012. 

Currently, he is pursuing a Ph.D.\ degree in the School of Electrical and Electronic Engineering, Nanyang Technological University, Singapore. From Aug-Dec 2015, he was a visiting student at Audiovisual Communication Laboratory (LCAV), \'{E}cole Polytechnique F\'{e}d\'{e}rale de Lausanne (EPFL), Switzerland. His research interests include remote sensing, statistical image processing and machine learning. 
\end{IEEEbiography}

\begin{IEEEbiography}
[{\includegraphics[width=1in,height=1.25in,clip,keepaspectratio]{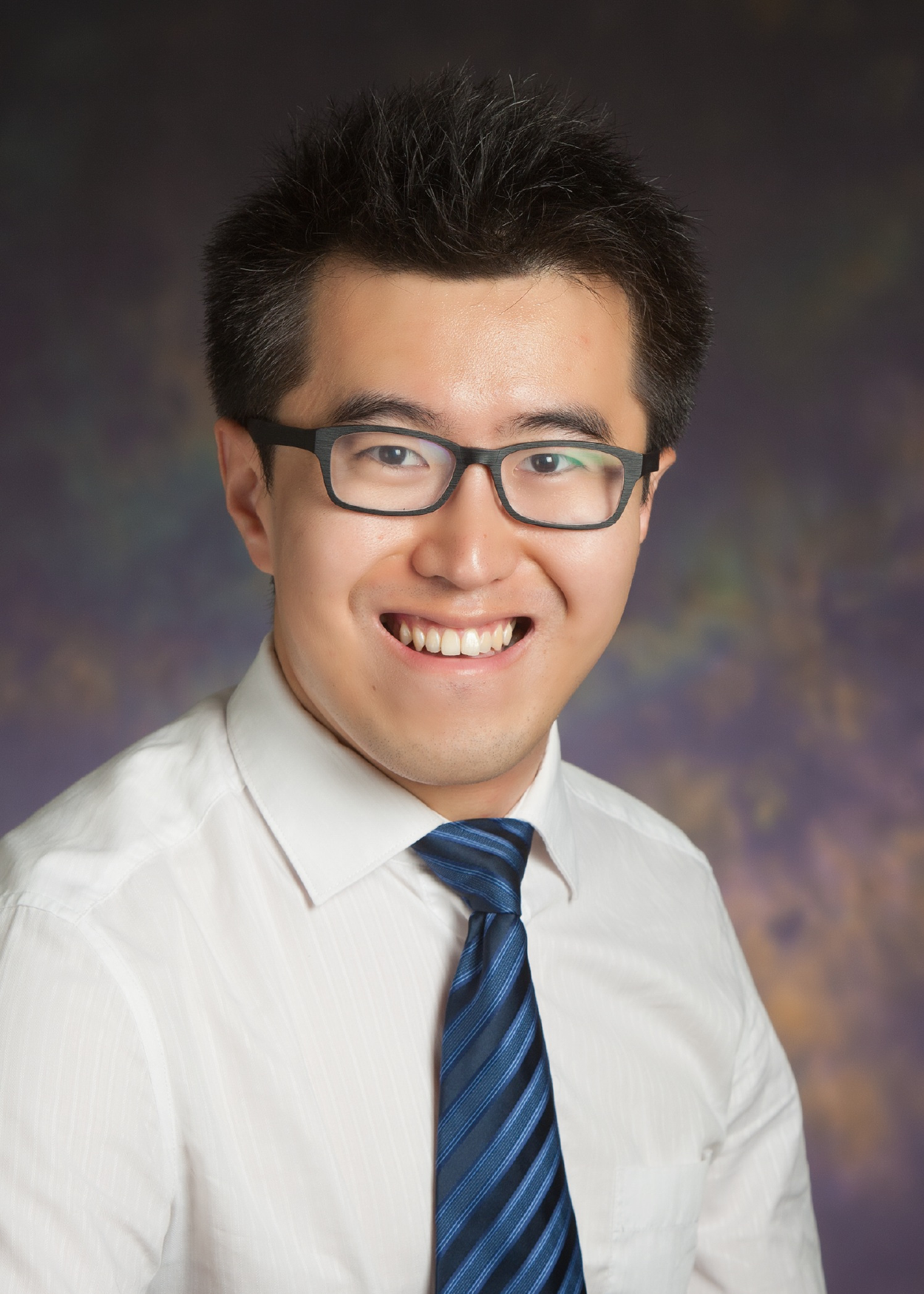}}]{Bihan Wen} received the B.Eng.\ degree in electrical and electronic engineering from Nanyang Technological University, Singapore, in 2012 and the M.S.\ degree in electrical and computer engineering from the University of Illinois at Urbana-Champaign, in 2015. 

He is currently pursuing the Ph.D. degree at the University of Illinois at Urbana-Champaign, Urbana, IL, USA. His current research interests include signal and image processing, machine learning, sparse representation, and big data applications.
	
\end{IEEEbiography}

\begin{IEEEbiography}
[{\includegraphics[width=1in,height=1.25in,clip,keepaspectratio]{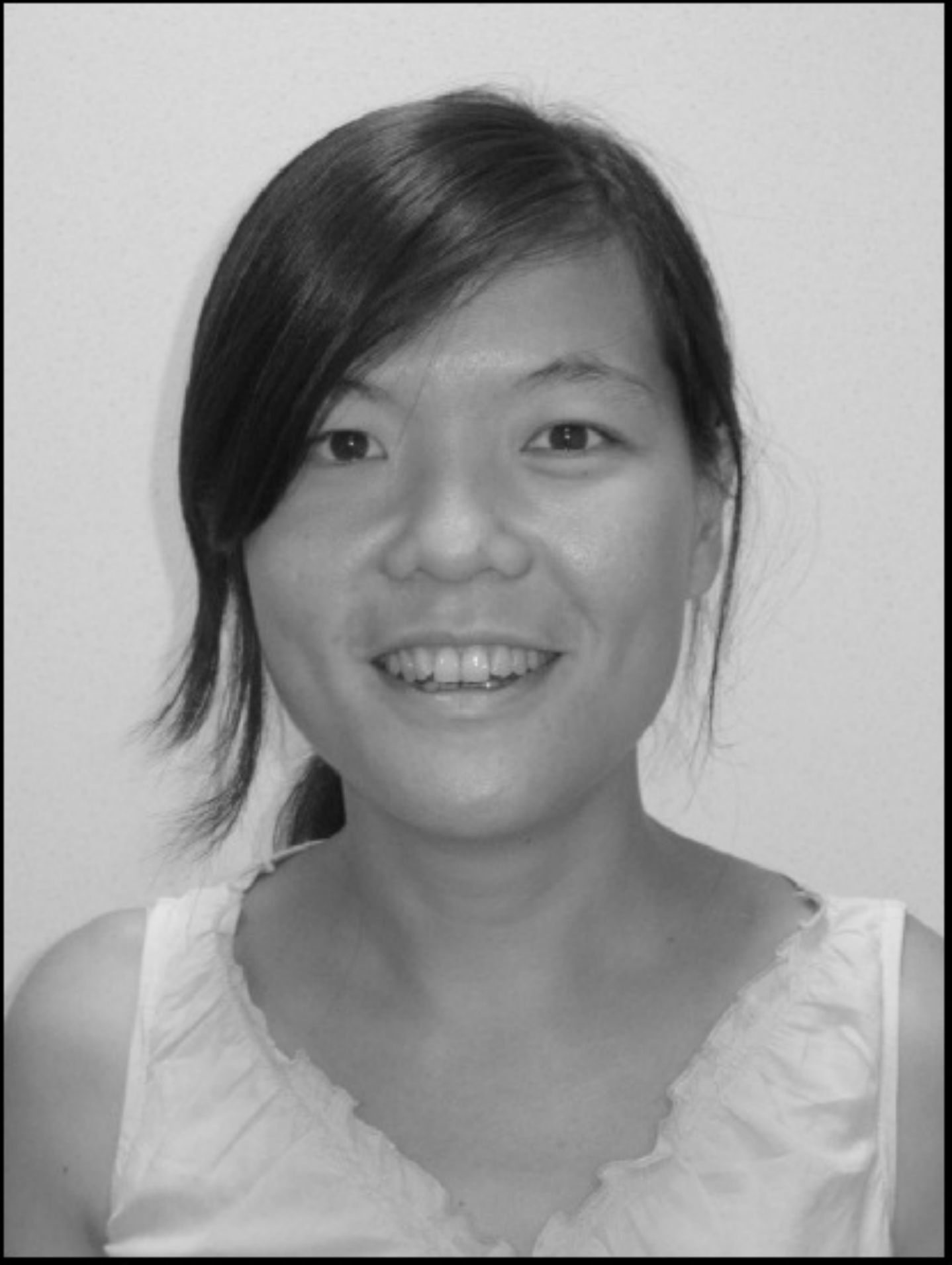}}]{Yee Hui Lee} (S'96-M'02-SM'11) received the B.Eng.\ (Hons.) and M.Eng.\ degrees from the School of Electrical and Electronics Engineering at  Nanyang Technological University, Singapore, in 1996 and 1998, respectively, and the Ph.D.\ degree from the University of York, UK, in 2002. 

Dr.\ Lee is currently Associate Professor and Assistant Chair (Students) at the School of Electrical and Electronic Engineering, Nanyang Technological University, where she has been a faculty member since 2002. Her interests are channel characterization, rain propagation, antenna design, electromagnetic bandgap structures, and evolutionary techniques.
\end{IEEEbiography}

\begin{IEEEbiography}
[{\includegraphics[width=1in,height=1.25in,clip,keepaspectratio]{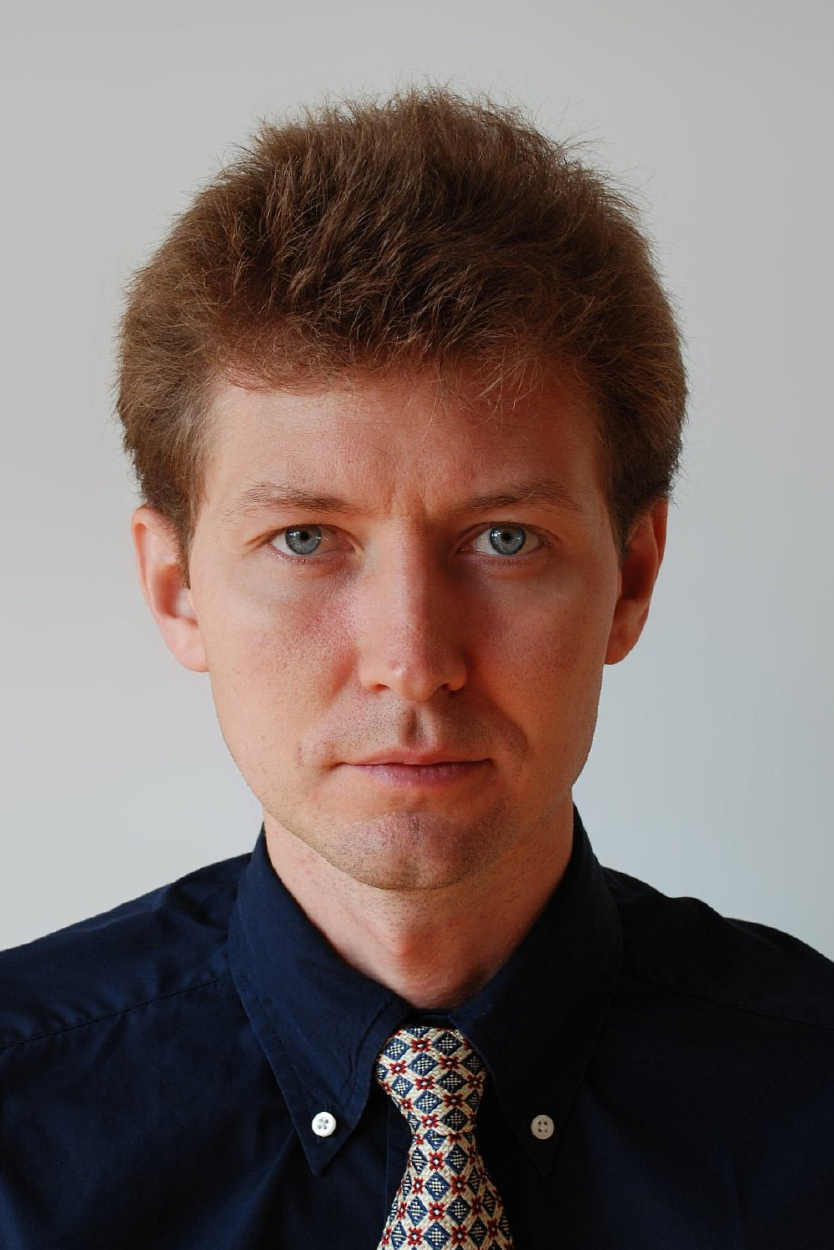}}]{Stefan Winkler} is Principal Scientist and Director of the Video \& Analytics Program at the University of Illinois' Advanced Digital Sciences Center (ADSC) in Singapore. Prior to that, he co-founded Genista, worked in several large corporations in Europe and the USA, and held faculty positions at the National University of Singapore and the University of Lausanne, Switzerland.

Dr.\ Winkler has a Ph.D.\ degree from the \'{E}cole Polytechnique F\'{e}d\'{e}rale de Lausanne (EPFL), Switzerland, and an M.Eng./B.Eng.\ degree from the University of Technology Vienna, Austria. He has published over 100 papers and the book ``Digital Video Quality'' (Wiley). He is an Associate Editor of the IEEE Transactions on Image Processing, a member of the IVMSP Technical Committee of the IEEE Signal Processing Society, and Chair of the IEEE Singapore Signal Processing Chapter. His research interests include video processing, computer vision, perception, and human-computer interaction.
	
\end{IEEEbiography}

\balance

\end{document}